\pdfoutput=1

\documentclass[11pt]{article}

\usepackage[preprint]{acl}

\usepackage{times}
\usepackage{latexsym}
\usepackage{booktabs}
\usepackage{booktabs,tabularx,makecell}
\usepackage{graphicx}
\usepackage{amssymb}
\usepackage{hyperref}
\usepackage{float}
\usepackage{arydshln} 
\usepackage{graphicx}
\usepackage{subcaption}
\usepackage{multirow}
\usepackage{graphicx, array, booktabs}
\newcolumntype{C}[1]{>{\centering\arraybackslash}m{#1}} 

\usepackage{lipsum} 

\usepackage{xltabular}

\usepackage[most]{tcolorbox}
\usepackage{xcolor}

\usepackage{soul}

\usepackage{amsmath}
\usepackage[T1]{fontenc}

\usepackage[utf8]{inputenc}

\usepackage{microtype}

\usepackage{inconsolata}

\usepackage{graphicx}

\usepackage{xcolor}

\title{DF-RAG: Query-Aware Diversity for Retrieval-Augmented Generation}

\author{
  \textbf{Saadat Hasan Khan\textsuperscript{ 1
\thanks{Majority of work done during internship at Capital One}}}  
  \textbf{Spencer Hong\textsuperscript{ 2}}  
  \textbf{Jingyu Wu\textsuperscript{ 2}}  
  \textbf{Kevin Lybarger\textsuperscript{ 1}}  \\
  \textbf{Youbing Yin\textsuperscript{ 2}}  
  \textbf{Erin Babinsky\textsuperscript{ 2}}   
  \textbf{Daben Liu\textsuperscript{ 2}}  
\\
  George Mason University\textsuperscript{1}  
  Capital One\textsuperscript{2} \\
  \texttt{\{skhan225, klybarge\}@gmu.edu} \\
  \texttt{\{spencer.hong, jingyu.wu, youbing.yin, erin.babinksy, daben.liu\}@capitalone.com} \\
\\
}

\begin{document}
\maketitle

\begin{abstract}


Retrieval-augmented generation (RAG) is a common technique for grounding language model outputs in domain-specific information. However, RAG is often challenged by reasoning-intensive question-answering (QA), since common retrieval methods like cosine similarity maximize relevance at the cost of introducing redundant content, which can reduce information recall. To address this, we introduce \textbf{Diversity-Focused Retrieval-Augmented Generation (DF-RAG)} that systematically incorporates diversity into the retrieval step to improve performance on complex, reasoning-intensive QA benchmarks. DF-RAG builds upon the Maximal Marginal Relevance framework to select information chunks that are both relevant to the query and maximally dissimilar from each other. A key innovation of DF-RAG is its ability to optimize the level of diversity for each query dynamically at test time without requiring any additional fine-tuning or prior information. We show that DF-RAG improves $F_{1}$ performance on reasoning-intensive QA benchmarks by 4–10\% over vanilla RAG using cosine similarity and also outperforms other established baselines. Furthermore, we estimate an Oracle ceiling of up to 18\% absolute $F_{1}$ gains over vanilla RAG, of which DF-RAG captures up to 91.3\%. 
\end{abstract}

\section{Introduction}

Retrieval-Augmented Generation (RAG) \cite{lewis2020rag, pmlr-v119-guu20a} is a common technique for grounding Large Language Models' (LLMs) outputs in external knowledge due to its lightweight and composable architecture. However, RAG is often challenged by complex, reasoning-intensive question answering (QA) tasks as these questions involve retrieving and integrating relevant information scattered across multiple contexts \cite{li2025structrag, liu2024raghelpreasoningllm}. One central limitation of common RAG implementations is its reliance on cosine similarity, which frequently over-selects semantically redundant content \cite{carbonell1998mmr, liu-etal-2025-pointwise} and excludes complementary evidence that is crucial for performing complex reasoning \cite{asai2024selfrag, xiong2021answering}. 

Fundamentally, for such complex, reasoning-intensive QA, a RAG system should only need to retrieve the few critical pieces of information required to bridge reasoning gaps to generate the right answer \cite{khattab}. We are thus motivated to ask our first research question: \\ \\
\textit{Can we improve our retrieval strategy to incorporate complementary evidence that current RAG approaches ignore such that it bridges reasoning gaps in complex QA tasks?}
\\

To answer this, we investigate the role of retrieval diversity in RAG pipelines for reasoning-intensive QA. We build upon Maximal Marginal Relevance (MMR) \cite{carbonell1998mmr}, a technique that balances query relevance with dissimilarity from selected information via a diversity parameter, $\lambda$. We hypothesize that strategically controlling retrieval diversity can bridge critical reasoning gaps required in complex reasoning QA and outperform current RAG baselines. 

We first test this hypothesis by replacing cosine similarity with MMR using a fixed $\lambda$ for all queries in a dataset. We find that the optimal $\lambda$ varies across datasets, which suggests that individual queries within a dataset may benefit from tailored $\lambda$ values. To estimate the upper bound of performance of dynamically selecting $\lambda$ at the query-level, we design an Oracle that selects the optimal  $\lambda$ for each query using the ground truth. The performance of our Oracle helps us
address our first research question: Oracle surpasses vanilla RAG by $\sim$18\% and also outperforms full-context baselines while utilizing $\sim$15\% of the context. However, estimating the right $\lambda$ at test time without any task-specific tuning is challenging. This leads to our second research question: \\ \\
\textit{How can we design a RAG system capable of incorporating the optimal level of diversity at test time?}
\\
\\
\indent To this end, we introduce \textbf{Diversity-Focused Retrieval-Augmented Generation (DF-RAG)}, a RAG pipeline designed to approximate the Oracle's performance without access to the ground truth. DF-RAG adaptively determines the optimal $\lambda$ for each query at test time through a multi-step process: (1) decomposing each query into individual steps with a planner LLM, (2) retrieving multiple candidate sets of chunks, each generated at predefined $\lambda$ levels; and (3) selecting the chunk set with the highest potential to address the decomposed steps with an Evaluator LLM. DF-RAG does not require fine-tuning, utilizes a plug-and-play retrieval framework, and directly addresses our second question. By benchmarking DF-RAG across five QA datasets, we demonstrate that leveraging DF-RAG to find the optimal $\lambda$ at test time can close the performance gap to the Oracle. \\
\indent To empirically validate our systems for such complex reasoning-intensive QA tasks, we benchmark them on: 1) multi-hop QA datasets as they require reasoning over evidence scattered across multiple contexts and 2) non-multi-hop but challenging long context QA datasets as they still require reasoning over context to generate answers \cite{yang-etal-2018-hotpotqa,xanh2020_2wikimultihop}. Our empirical evaluation  across $\sim$180 experiments shows that DF-RAG routinely outperforms strong baselines on multi-hop QA, achieving up to 9.5\% gains over vanilla RAG across context lengths and closing as much as up to 91.3\% of the gap to the Oracle. On non-multi-hop QA, DF-RAG achieves up to a 4\% improvement over vanilla RAG and other baselines. 
To our knowledge, this is the first work to leverage query-level diversity to estimate Oracle performance bounds and to propose a  RAG framework that narrows this gap by dynamically balancing diversity and relevance at test time. 
\section{Related Work}
\subsection{Multi-hop Question Answering}

\begin{figure*}[h]
    \centering
    \includegraphics[width=\textwidth]{ 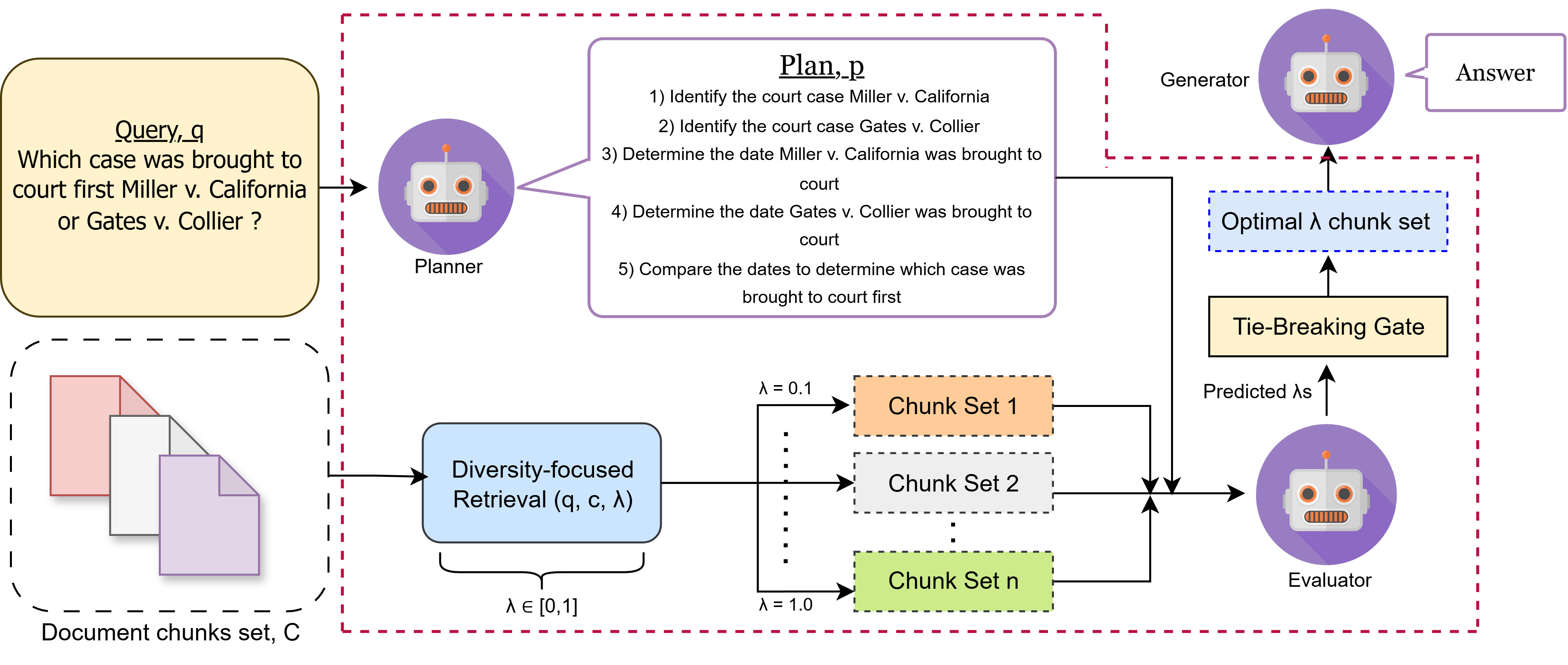}
    \caption{Architecture of DF-RAG: A training-free framework that dynamically selects the right level of diversity to generate answers for reasoning-intensive QA tasks. DF-RAG adaptively adjusts diversity for each query by utilizing the Planner, Evaluator and the Diversity-focused Retrieval. We illustrate this with an example query from HotpotQA \cite{yang-etal-2018-hotpotqa}.}
    \label{fig:fullwidth_diagram}
\end{figure*}

Early work on multi-hop QA established foundational benchmarks like HotpotQA \cite{yang-etal-2018-hotpotqa} and MuSiQue \cite{trivedi-etal-2022-MuSiQue}, which require finding and reasoning over multiple supporting documents. Many methods perform the task by conducting inference over static or dynamic graphs to find reasoning paths \cite{fang-etal-2020-hierarchical, ding-etal-2019-cognitive, 10.1145/3404835.3462853}. Neural-based approaches have also attempted to decompose multi-hop questions into single-hop questions or generate follow-up questions based on retrieved information \cite{min-etal-2019-multi, cao-etal-2019-bag, malon2020generatingfollowupquestionsinterpretable}. Recent advances in multi-hop QA include LLM-based approaches that leverage strong natural language understanding capabilities \cite{xu-etal-2021-exploiting-reasoning} through reasoning chain generation \cite{reasoning_chain}, question decomposition \cite{wu2024gendecrobustgenerativequestiondecomposition}, and graph construction \cite{li-du-2023-leveraging}. 

\subsection{Retrieval Augmented Generation}

Despite the superior language understanding and generation capabilities of LLMs, their performance remains limited on various downstream tasks due to challenges, such as outdated and long-tail knowledge \cite{he2022rethinking, pmlr-v202-kandpal23a}, hallucination \cite{10.1145/3624918.3625329} and the lack of domain-specific knowledge \cite{shen2023chatgpt}. RAG \cite{pmlr-v119-guu20a,lewis2020rag} has become the most widely adopted approach to address these issues by grounding LLM outputs in external knowledge for domain-specific downstream tasks. RAG pipelines operate by retrieving the most relevant document segments for a given query, and their effectiveness is therefore strongly influenced by the chunking strategies employed during retrieval. Recent advances in RAG include methods for reducing noise from chunks \cite{yan2024crag, zhuang-etal-2023-open}, chunk-free strategies to mitigate semantic loss \cite{qian2024cfic}, hierarchical arrangements of chunks via summarization techniques to capture global context \cite{sarthi2024raptor}, adaptive chunk-length strategies \cite{li-etal-2024-retrieval}, and approaches that extract global cues to generate supporting questions for improved answers \cite{zhao-etal-2024-longrag}.  

\subsection{Diversity in Information Retrieval}

Traditionally, information retrieval has primarily emphasized relevance. Early work  introduced Maximal Marginal Relevance (MMR) \cite{carbonell1998mmr}, which balances relevance and novelty. This inspired intent-aware ranking models aimed at improving subtopic coverage and evaluation \citep{agrawal2009diversifying, clarke2008novelty}. Some approaches have incorporated explicit diversification objectives into learning-to-rank frameworks, jointly optimizing for relevance and subtopic coverage \citep{santos2015search}. 
Recent efforts have also explored diversity in RAG-based pipeline to reduce redundant evidence for downstream taks \cite{zhang-etal-2025-lexical, wang2025diversityenhancesllmsperformance}. While these RAG pipelines highlight the importance of diversity, they either depend on additional fine-tuning or lack mechanisms to adaptively determine the optimal diversity, limiting their generalizability and effectiveness in practice.

\section{Methodology}

\subsection{Diversity-focused Retrieval}
\label{sec:scoring function}

To reduce content redundancy, we design a diversity-focused retrieval function 
built on MMR \cite{carbonell1998mmr}. Given a query $q$, a candidate chunk $c \in \mathcal{C}$, and a set of previously 
selected chunks $S = \{s_1, s_2, \dots\}$, the retrieval function assigns a geometric MMR score (gMMR) to $c$ as follows:


\newcommand{\cS}{\vec{c}_{S}} 

\begin{equation}
\label{eq:mmr_final}
\begin{split}
\text{gMMR}(c) &= \lambda\,\operatorname{cos}(\vec{q},\vec{c}) \\
&\quad + (1-\lambda)\,\sqrt{\,2 - 2\,\operatorname{cos}\!\left(\vec{c}, \cS\right)}
\end{split}
\end{equation}

Where the hyperparameter $\lambda$ modulates the trade-off between relevance and diversity, and $\cS$ is the centroid of $S$ so far, calculated by:

\begin{equation}
\label{eq:set_vector}
\vec{c}_S \triangleq \frac{1}{|S|} \sum_{s \in S} \vec{s}
\end{equation}

To measure diversity, we replace standard cosine similarity with the Euclidean distance between the normalized embeddings, calculated as $\sqrt{\,2 - 2\cos\!\left(\vec{c}, \cS\right)}$ . This approach offers two advantages. First, it grounds diversity in a geometric framework enabling us to frame diversity as the spatial distance between a candidate chunk ($\vec{c}$) and the centroid of the already selected chunks ($\cS$). Second, adopting this distance-based variant of MMR yields improved empirical performance in our downstream tasks. A detailed comparison against classical MMR is provided in Appendix \ref{sec:eucvs}.


We greedily select the chunk $c^*$ at the $k$-th step using the following conditions:
\begin{equation}
\label{eq:chunk_selection_piecewise}
c^* =
\left\{
\begin{aligned}
& \arg\max_{\,c \in \mathcal{C}} \cos(\vec{q}, \vec{c})
&& \text{if } S = \varnothing, \\
& \arg\max_{\,c \in \mathcal{C}\setminus S} \text{gMMR}(c)
&& \text{otherwise.}
\end{aligned}
\right.
\end{equation}

%


\subsection{Baseline: RAG with $\lambda$-Fixed gMMR}
\label{sec:RAGwMMR}
We implement a baseline using RAG with our gMMR function (Equation~\ref{eq:mmr_final}). In this approach, a constant $\lambda$ is applied uniformly to all queries in a given benchmark. We vary $\lambda \in [0,1]$ in increments of 0.1 to identify the value that yields the highest $F_1$ score for each dataset.

\subsection{Oracle: A Theoretical Upper Bound for Query-Adaptive gMMR}
We introduce an Oracle to represent the hypothetical upper bound for our query-adaptive approach. This Oracle is designed to determine the ideal gMMR diversity parameter $\lambda$  for each query by leveraging the ground truth. This is achieved by sampling values from 0 to 1 in increments of 0.1 and selecting the $\lambda$s whose corresponding chunk set yields the highest $F_1$ score with respect to the ground truth.

\subsection{The DF-RAG Architecture}
\label{sec:df-rag architecture}
The Oracle establishes a theoretical performance upper bound for our query-adaptive gMMR. We next introduce DF-RAG, an architecture designed to approximate Oracle at test time for each query. 
As depicted in Figure~\ref{fig:fullwidth_diagram}, the architecture of DF-RAG integrates a plug-and-play retrieval component (highlighted in the purple dashed box) that uses an LLM-based Planner and an Evaluator for selecting the optimal $\lambda$ for each query. 
When a query, \textit{q}, is received, two processes are initiated in parallel: plan generation via the Planner as well as candidate chunk set retrieval using gMMR. Once the plan, \textit{P}, and the candidate chunk sets are available, the Evaluator evaluates and selects the optimal chunk set thereby determining the optimal $\lambda$. Finally, given the selected chunk set, the Generator synthesizes a final answer to the query. These components are explained in more details below.

\noindent\textbf{Planner for Step Decomposition:} Given a query, \(q\), the Planner generates an explicit sequence of steps \(\{p_1, p_2, \dots  p_m\} \) that outline how the query should be solved, which we denote as a plan, \(P\). Our hypothesis is that any candidate chunk set \(S\) sufficient to answer \(q\) should also be sufficient to execute \(P\) with the collective information it holds. Therefore, we assess the quality of a given \(S\) based on how well it supports \(P\)  to make an informed prediction of \(\lambda\) at test time. The Planner operates under a one-shot prompting setup, with the prompting details detailed in Appendix \ref{sec:systemprompts}.

\noindent\textbf{Candidate Chunk Sets Retrieval with gMMR:} We retrieve candidate chunk sets,  $S = \{s_1,s_2,\dots,s_k\}$, from the document set, $\mathcal{C}$, using gMMR (Section \ref{sec:scoring function}). For our experiments, we sample chunks uniformly with $\lambda \in [0.1,1]$ with a step size of $0.1$. We exclude sampling a chunk set at $\lambda = 0$ because selecting chunks $2 \dots k$ at $\lambda = 0$ disregards query relevance and introduces noisy information. The computational complexity of DF-RAG depends directly on the sampling strategy and uniform sampling yields a $O(n)$ complexity. Additionally, we propose a simpler binary-search-based sampling strategy that achieves $O(\log(n))$ complexity with comparable performance. More details and corresponding results of binary-search based sampling strategy for DF-RAG is presented in Appendix \ref{sec:bs}.

\noindent\textbf{Evaluator for Optimal Diversity Selection:} Given a pool of candidate chunk sets retrieved using different $\lambda$s, and a plan by the Planner, \(P=\{p_i\}_{i=1}^m\), for the query, $q$, the Evaluator evaluates each candidate chunk set with respect to $P$. For each item $p_i \in P$, the Evaluator assigns a support score \(r(p_i)\in\{0,\dots,5\}\), where \(0\) indicates no supporting information is present and \(5\) indicates direct evidence satisfying that step in the candidate chunk set. The overall score, denoted by $Retreival\ Support$, of a chunk set retrieved under \(\lambda\) is then computed as:
\begin{equation}
Retreival\ Support(\lambda, P) = \sum_{i=1}^m r(p_i)
\end{equation}
The optimal query-specific $\lambda$, denoted $\lambda^*$, is then selected as follows:
\begin{equation}
\lambda^* = \arg\max_{\lambda \in \Lambda} Retreival \ Support(\lambda, P)
\end{equation}

The Evaluator utilizes a few-shot prompting approach and the prompting details and the specific few-shot examples are discussed in Appendix \ref{sec:systemprompts}.


\noindent\textbf{Tie-Breaking Gate:} DF-RAG's Evaluator's assessment may identify multiple \(\lambda\)s as equally optimal for a given query, but the system must select a single \(\lambda\) and its corresponding chunk set for answer generation due to the limit on context length we impose. To handle this, we introduce a Tie-Breaking Gate that selects one chunk set from the set of tied candidates. Our tie-breaking mechanism is deliberately lightweight and operates by selecting the median \(\lambda\) (upper median in case of even number of selected values) among the tied candidates. 

\noindent\textbf{Generator for Final Answer Synthesis:} Similar to standard RAG systems, the final component is an LLM-based Generator. It is tasked with producing an answer given the original query \textit{q} and the optimally-diverse chunk set  $S^*$ retrieved using  $\lambda^*$.  We follow the prompt released in LongBench \cite{bai-etal-2024-longbench} for answer generation, which is further detailed in Appendix \ref{sec:systemprompts}.

\begin{table*}[!t]
\centering
\resizebox{\textwidth}{!}{%
\begin{tabular}{lcccccccccc}
\toprule
\multirow{2}{*}{\textbf{Experiment Type}} &
\multicolumn{5}{c}{\textbf{Llama 3.3 70B}} &
\multicolumn{5}{c}{\textbf{Qwen 2.5 72B}} \\
\cmidrule(lr){2-6} \cmidrule(lr){7-11}
& \textbf{HotpotQA} & \textbf{MuSiQue} & \textbf{2WikiMQA} & \textbf{MultifieldQA} & \textbf{En.QA}
& \textbf{HotpotQA} & \textbf{MuSiQue} & \textbf{2WikiMQA} & \textbf{MultifieldQA} & \textbf{En.QA} \\
\midrule



\multicolumn{11}{c}{\textit{context length = 500 words }} \\
\midrule
Oracle & 69.5 & 46.2 & 65.3 & 54.9 & 41.3 & 67.7 & 43.8 & 64.8 & 57.2 & 40.1 \\
\hdashline
Vanilla RAG & 51.1 & 29.9 & 49.8 & 47.7 & 26.3 & 51.2 & 24.8 & 47.7 & 47.5 & 26.1\\
RAG with gMMR  & 53.3 & 30.9 & 54.8 & \underline{48.2} & \underline{27.6} & 55.0 & 27.6 & 52.7 & 49.7 & \underline{27.7}\\
DF-RAG  & \textbf{58.1} & \underline{33.3} & \textbf{58.1} & \textbf{48.3} & \textbf{29.3} & \underline{58.5} & \underline{31.3} & \underline{53.7} & \textbf{51.0} & \textbf{28.2} \\
DF-RAG  + IC & \underline{56.8} & \textbf{34.0} & \underline{55.2} & 47.0 & 27.5 & \textbf{60.1} & \textbf{32.7} & \textbf{56.8} & \underline{48.8} & \underline{27.7} \\
\midrule

\multicolumn{11}{c}{\textit{context length = 1000 words}} \\
\midrule
Oracle & 68.6 & 49.7 & 68.6 & 56.2 & 45.2 & 67.5 & 48.7 & 67.2 & 56.7 & 44.1 \\
\hdashline
Vanilla RAG & 56.0 & 36.0 & 58.2 & 50.0 & 28.3 & 55.9 & \underline{31.9} & 55.0 & \underline{50.6} & 26.9 \\
RAG with gMMR  & 56.0 & \underline{37.5} & 58.7 & \underline{50.9} & \underline{30.1} & 56.7 & \textbf{33.3} & 55.5 & \textbf{51.9} & \underline{30.3}\\
DF-RAG & \textbf{60.3} & 36.3 & \underline{58.9} & \textbf{51.6} & \textbf{31.5} & \underline{60.4} & 28.8 & \underline{57.4} & 50.4 & \textbf{31.1} \\
DF-RAG + IC & \underline{58.4} & \textbf{38.6} & \textbf{63.3} & 48.6 & 29.3 & \textbf{62.1} & 31.7 & \textbf{59.7} & 49.3 & 29.6 \\
\midrule

\multicolumn{11}{c}{\textit{context length = 1500 words}} \\
\midrule
Oracle & 69.5 & 51.9 & 74.9 & 56.1 & 47.8 & 68.7 & 48.6 & 69.9 & 55.9 & 47.4 \\
\hdashline
Vanilla RAG & 56.1 & 35.4 & 61.6 & 50.4 & \underline{33.1} & 57.9 & \underline{33.6} & 59.5 & 48.1 & 31.3 \\
RAG with gMMR  & 58.1 & 35.9 & 63.4 & \textbf{50.9} & \textbf{33.2} & 60.1 & \textbf{34.6} & 59.8 & \textbf{50.1} & 32.5\\ 
DF-RAG & \underline{59.6} & \underline{37.9} & \underline{63.9} & \underline{50.6} & \underline{33.1} & \textbf{61.8} & 32.2 & \underline{62.1} & \underline{50.0} & \textbf{34.0} \\
DF-RAG + IC & \textbf{61.8} & \textbf{38.0} & \textbf{68.6} & 46.5 & 31.8 & \underline{60.9} & 33.3 & \textbf{69.0} & 48.7 & \underline{33.2} \\
\midrule

\multicolumn{11}{c}{\textit{context length independent baselines}} \\
\midrule
LongRAG & 57.4 & 38.1 & 61.7 & \textbf{52.9} & 22.9 & 57.7 & 36.8 & 59.9 & 47.8 & 22.4 \\
RAPTOR & 54.5 & 34.3 & 58.9 & 48.3 & X & 53.0 & 29.6 & 58.4 & 49.6 & X \\
HippoRAG & 46.2& 33.5 & 57.1 & 48.3 & X & 47.0 & 32.8 & 53.2 & 48.9 & X \\
\bottomrule
\end{tabular}
}
\caption{$F_1$ scores of vanilla RAG and DF-RAG on five QA datasets across varying context lengths, evaluated using Llama 3.3 70B and Qwen 2.5 72B. Context length is varied by changing chunk size $w \in \{100,200,300\}$ words. \textbf{Bold} and \underline{underlined} results indicate first and second in $F_1$ score. Results marked with X could not be performed due to memory limit restrictions.}
\label{tab:performance_acl_dfrag_transposed}
\end{table*}

\subsection{Experimental Details}
\noindent\textbf{Preliminaries:} We perform comprehensive experiments using Qwen 2.5 72B \cite{qwen2025qwen25technicalreport} and Llama 3.3 70B \cite{grattafiori2024llama3herdmodels} as backbone LLMs. We use 16-bit floating precision and run our experiments on 2 NVIDIA A100 80GB GPUs. We perform retrieval based on sentence embeddings generated by `multi-qa-mpnet-base-cos-v1'. For our proposed experiments (other than established baselines), we always retrieve 5 chunks where chunks are generated at the word level. We vary context lengths by varying the number of words, \(w \in \{100, 200, 300\}\), corresponding to context lengths of approximately 500, 1000, and 1500 words. All our results are based on single-run experiments. \\ 

\noindent\textbf{RAG with $\lambda$-Fixed gMMR:} We benchmark RAG with gMMR against vanilla RAG to highlight the impact of diversity-focused retrieval. Both models use the same generator and differ only in their chunk selection strategy. Vanilla RAG selects the top 5 chunks based on cosine similarity of their embeddings, whereas RAG with $\lambda$-Fixed gMMR selects chunks using gMMR at fixed $\lambda$ values across datasets.

\noindent\textbf{Oracle Upper Bound:} We benchmark our Oracle against vanilla RAG, RAG with $\lambda$-Fixed gMMR, and a long-context baseline. The long-context baseline (denoted as Long Context) is implemented by concatenating the entire context with the input query and feeding it to the LLM generator. This baseline is only applicable when the full context fits within the model’s token limit, which excludes En.QA from long-context evaluation.

\noindent\textbf{DF-RAG Baselines:} 
We implement two DF-RAG variants: (1) DF-RAG , where median \(\lambda\) is the tie-breaking gate used and (2) DF-RAG + incremental context, denoted by IC. The IC serves as a condensed summary added to the Generator's context. The IC comprises the plan produced by the Planner and the short answer generated by the Evaluator when executing that plan. The motivation and analysis of ICs benefits are further detailed in Appendix \ref{sec:ic}. We compare our DF-RAG variants to the Oracle and five baselines. The five baselines are: RAG with $\lambda$-Fixed gMMR, Vanilla RAG, RAPTOR \cite{sarthi2024raptor}, LongRAG \cite{zhao-etal-2024-longrag} and HippoRAG \cite{hipporag}. To experiment with RAPTOR, LongRAG and HippoRAG, we replace the backbone LLM while keeping all other implementation details consistent with their published setups. For LongRAG, we employ both the filter and extractor modules, as this configuration yields the best performance in their reported results. For HippoRAG, we use both Llama 3.3 70B and Qwen 2.5 72B for answer generation to be consistent with other baselines, however, the entity extraction phase was carried out with a Llama 3.3 70B due to HippoRAG's reported issues on some models' compatibility at the time of writing this paper. We benchmark against LongRAG, RAPTOR and HippoRAG given their demonstrated success in prior work and their strategic designs for retrieving critical information in multi-hop queries. All three architectures excel at complex reasoning tasks, making them well-suited as baselines.

\noindent\textbf{Datasets:} We benchmark our baselines, Oracle and DF-RAG on multiple reasoning-intensive QA datasets. Following prior work \cite{zhao-etal-2024-longrag, li-etal-2024-retrieval}, we select datasets that (a) are in English, (b) contain queries that require reasoning, (c) are structured for QA evaluation. To this end, our benchmark suite includes three multi-hop QA datasets from LongBench \cite{bai-etal-2024-longbench} — HotpotQA, MuSiQue, and 2WikiMultihopQA — one non–multi-hop QA dataset from LongBench, MultifieldQA (English version), and one dataset from $\infty$Bench \cite{zhang-etal-2024-bench}, En.QA. Together, these datasets cover diverse domains, including Wikipedia articles, meetings, narratives, and research papers, making it challenging to reason over the context when answering the queries. Datasets' statistics are presented in Appendix \ref{dataset-stat}.

\noindent\textbf{Evaluation Metrics:} We follow the official metrics of LongBench and \(\infty\)Bench for the selected benchmarks, reporting answer-level $F_1$ scores against the ground truth.

\section{Experimental Results}

\subsection{RAG with $\lambda$-Fixed gMMR}
\label{sec:RAGwMMRres}
Our results revealed that RAG with $\lambda$-Fixed gMMR outperforms vanilla RAG (see Table \ref{tab:performance_acl_dfrag_transposed} and Figure \ref{fig:oracvslc}). We also found that numerous distinct dataset-level $\lambda$ values outperformed vanilla RAG (see Figure \ref{fig:RAGwMMR}). This helped to confirm that diversity-focused methods are beneficial, motivating our investigation into whether an optimal $\lambda$ could be determined at the query-level. Details of the performance trends for each of these benchmarks across $\lambda$ is further detailed in Appendix \ref{sec:RAGWMMRcontd}.

\subsection{Oracle: Upper Bound for Query-Adaptive gMMR}
 Our Oracle consistently outperforms both vanilla RAG and RAG with $\lambda$-Fixed gMMR across all benchmarks (see Figure \ref{fig:oracvslc}). When compared to vanilla RAG, we observe consistent gains of $\sim$7--19\% $F_1$ points with the largest gains observed for multi-hop QA benchmarks. A similar behavior is also observed as the Oracle beats RAG with $\lambda$-Fixed gMMR by $\sim6- 16\%$. Furthermore, the Oracle also outperforms Long Context by up to 9.2\%, even when using up to $\sim 70-87$\% less context (see Figure \ref{fig:oracvslc}).  
 This demonstrates that query-aware, optimally-tuned retrieval diversity can be more effective than processing the entire context, provided we find the right $\lambda$. As the Oracle achieves the highest performance among all baselines, it serves as DF-RAG’s performance target. Accordingly, a dashed line follows Oracle’s score in Table \ref{tab:performance_acl_dfrag_transposed}.

\begin{figure}[!t]
    \centering
    \includegraphics[width=0.9\linewidth]{ 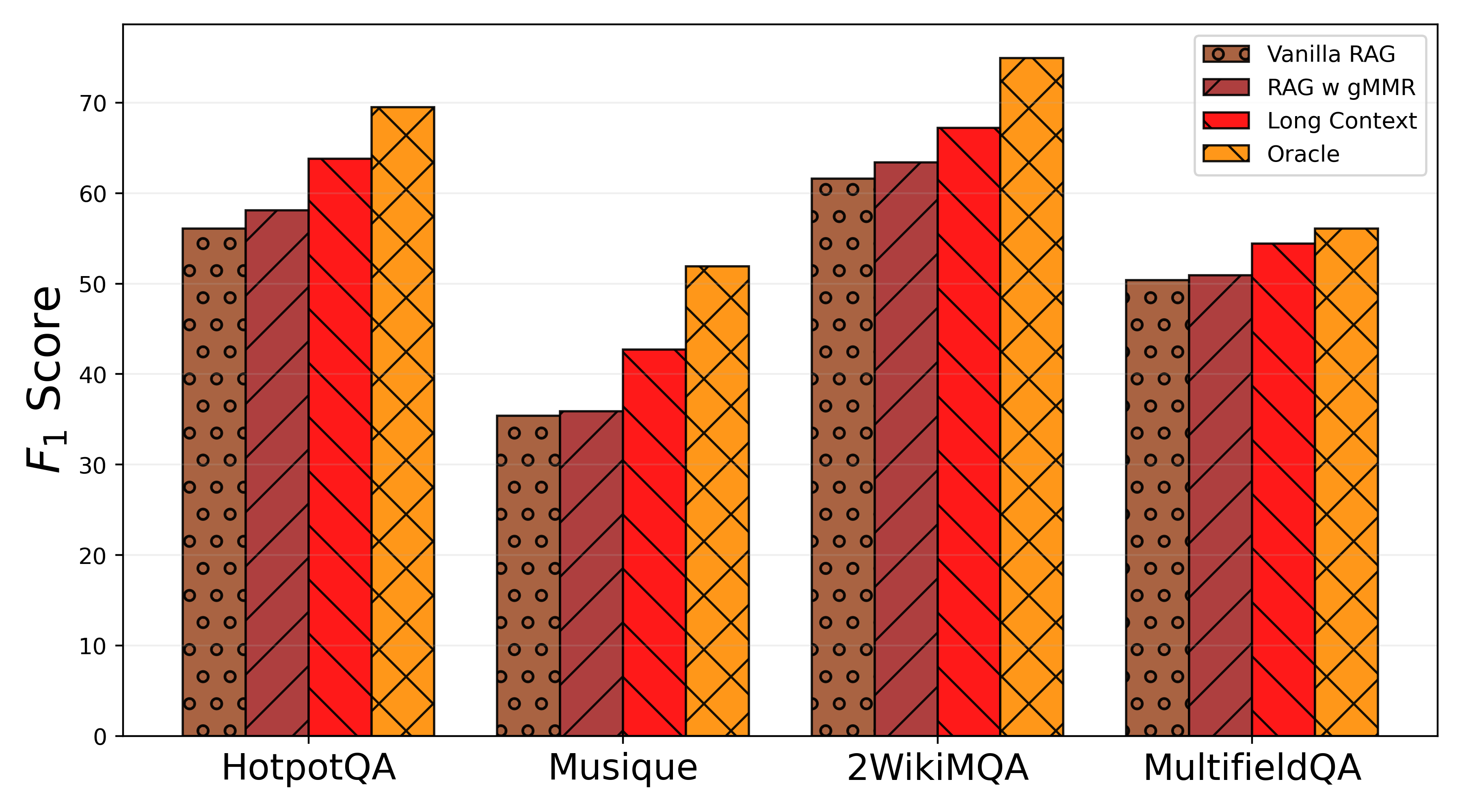}
    \caption{$F_1$ scores of vanilla RAG, RAG with gMMR, Long Context, and Oracle across LongBench benchmarks. Vanilla RAG, RAG with gMMR and Oracle operate at 1500 word context length. Results on En.QA are not reported as its context length exceeds our context limit.  }
    \label{fig:oracvslc}
\end{figure}

\label{sec:roacle_results}

\begin{figure*}[h]
  \centering
  \begin{subfigure}[t]{0.19\linewidth}
    \includegraphics[width=\linewidth]{ 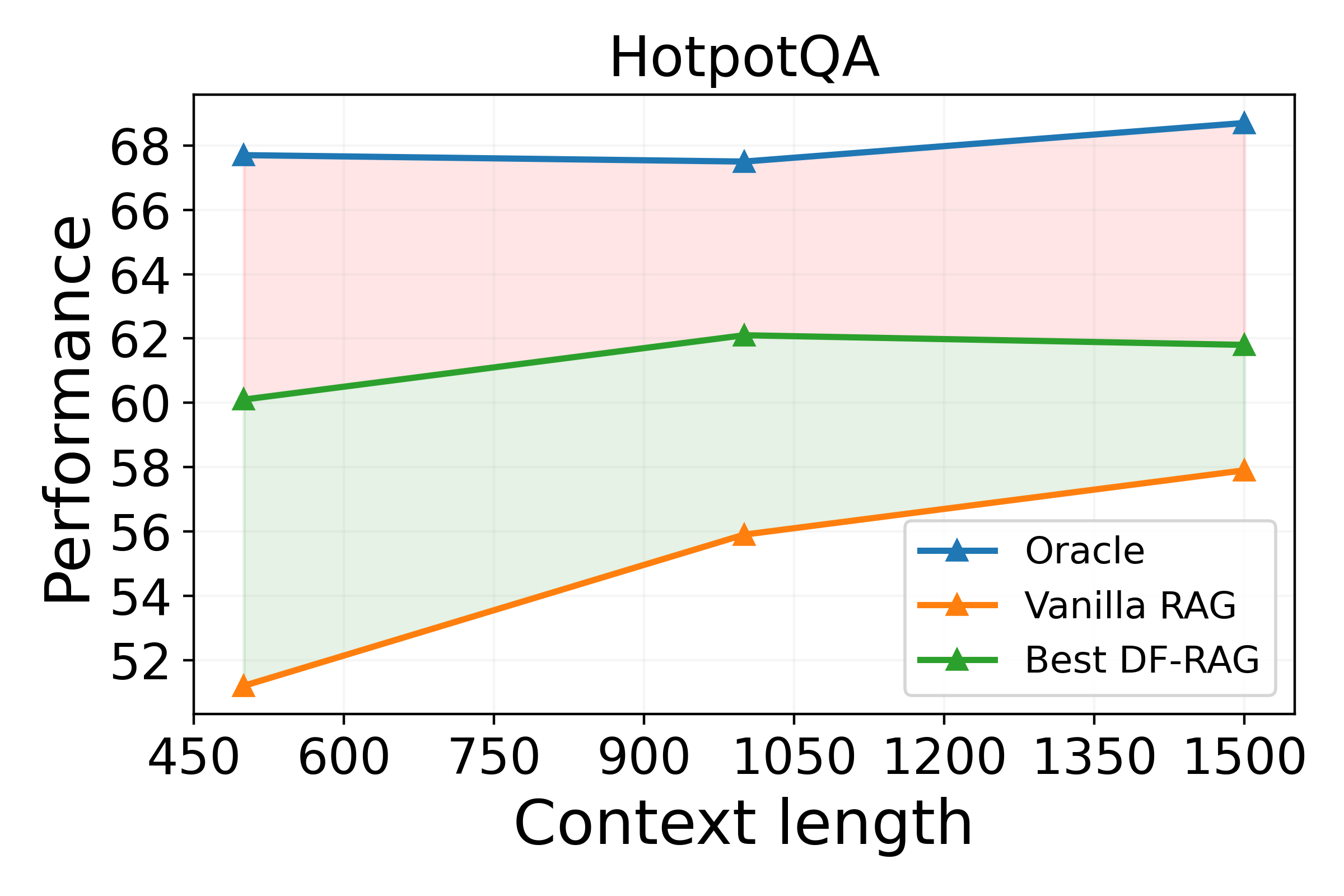}
  \end{subfigure}\hfill
  \begin{subfigure}[t]{0.19\linewidth}
   \includegraphics[width=\linewidth]{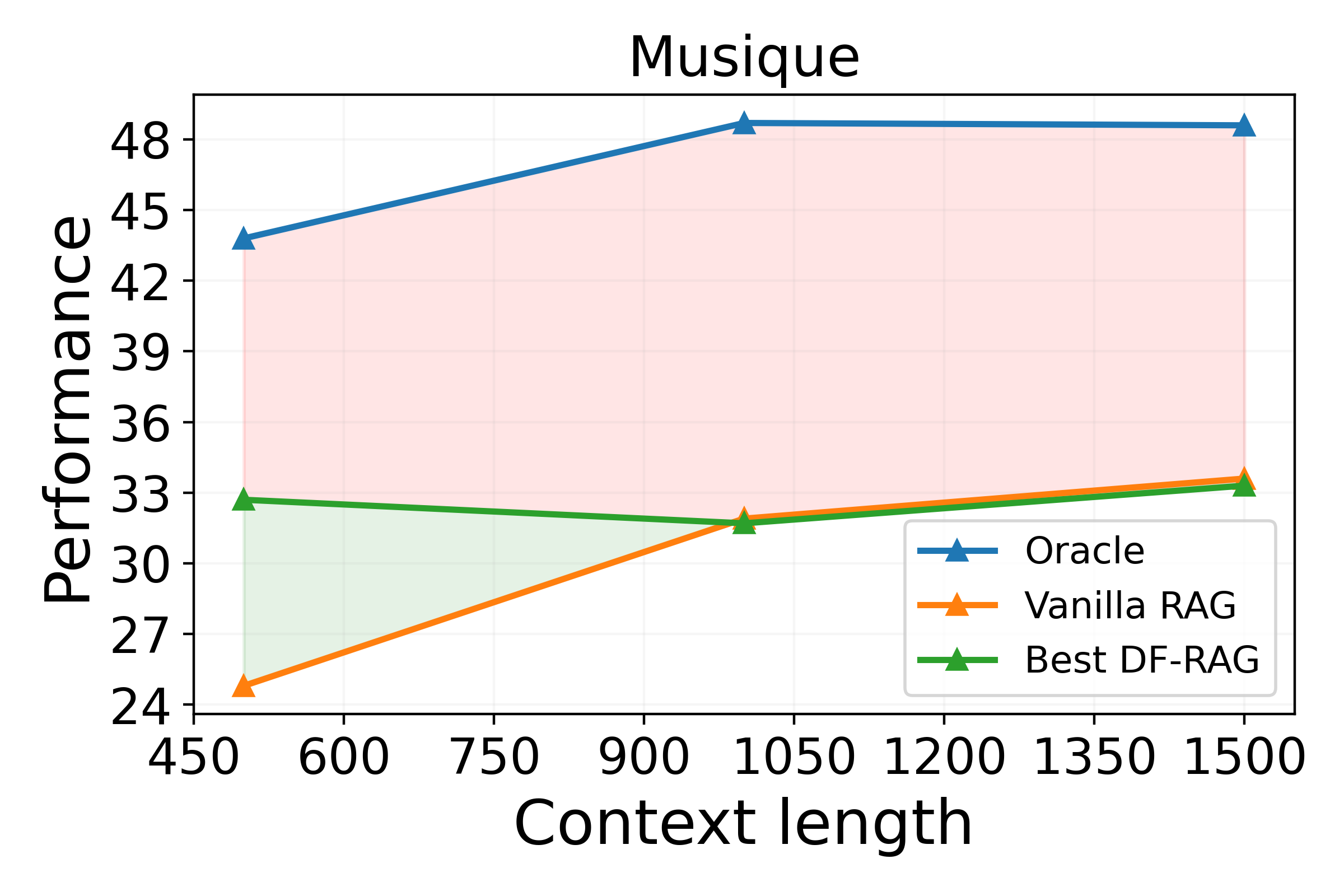}
  \end{subfigure}\hfill
  \begin{subfigure}[t]{0.19\linewidth}
    \includegraphics[width=\linewidth]{ 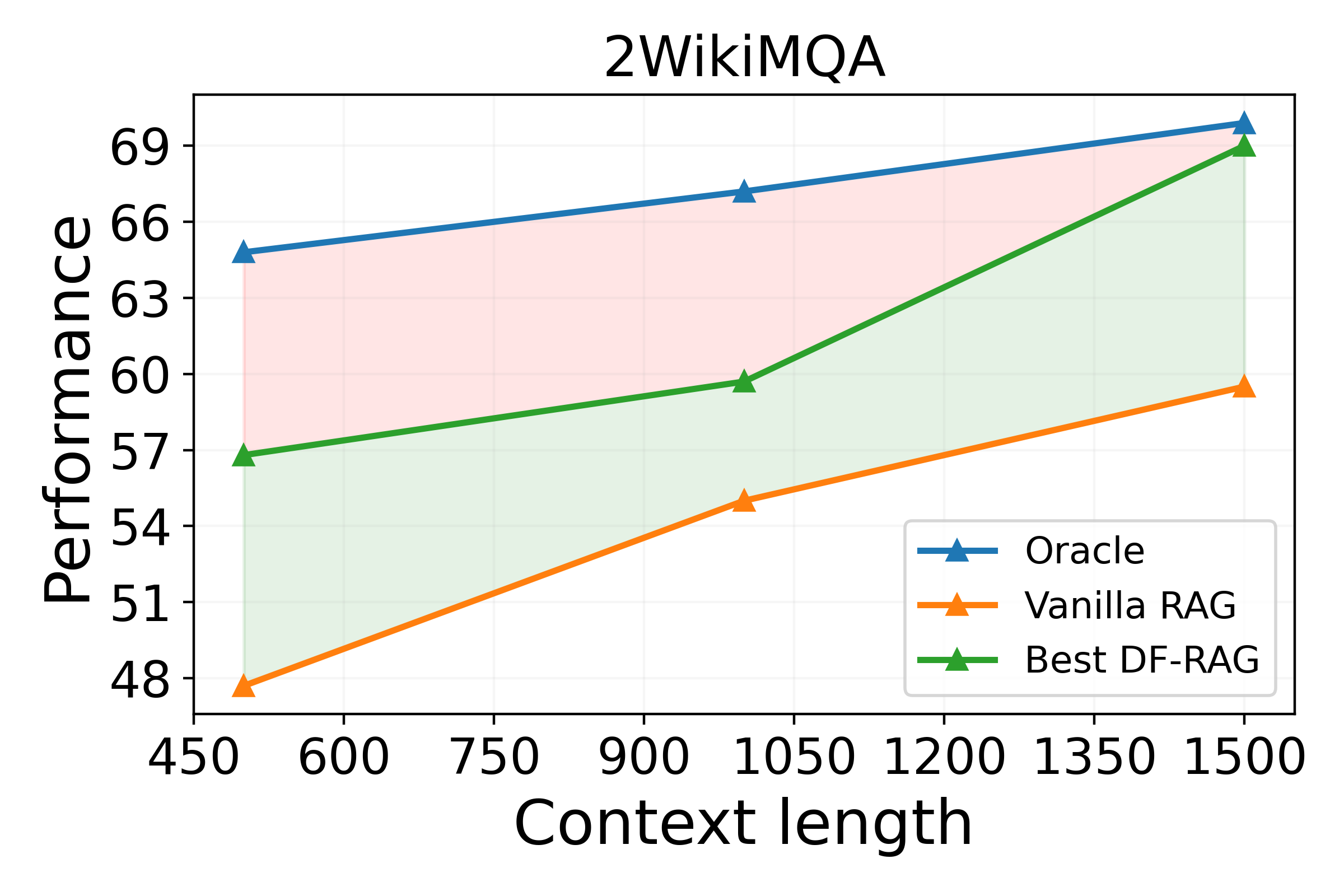}
  \end{subfigure}\hfill
  \begin{subfigure}[t]{0.19\linewidth}
    \includegraphics[width=\linewidth]{ 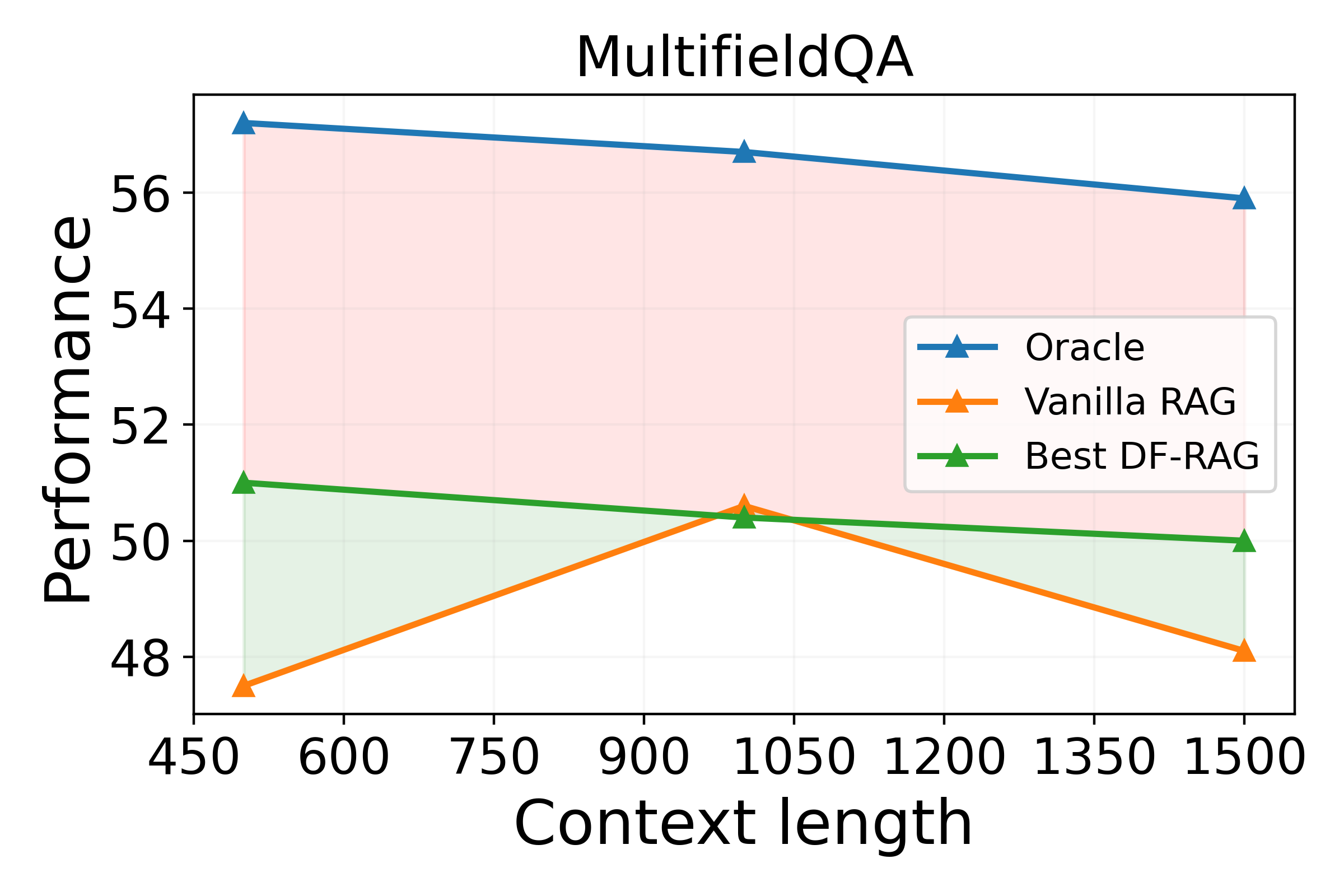}
  \end{subfigure}\hfill
  \begin{subfigure}[t]{0.19\linewidth}
    \includegraphics[width=\linewidth]{ 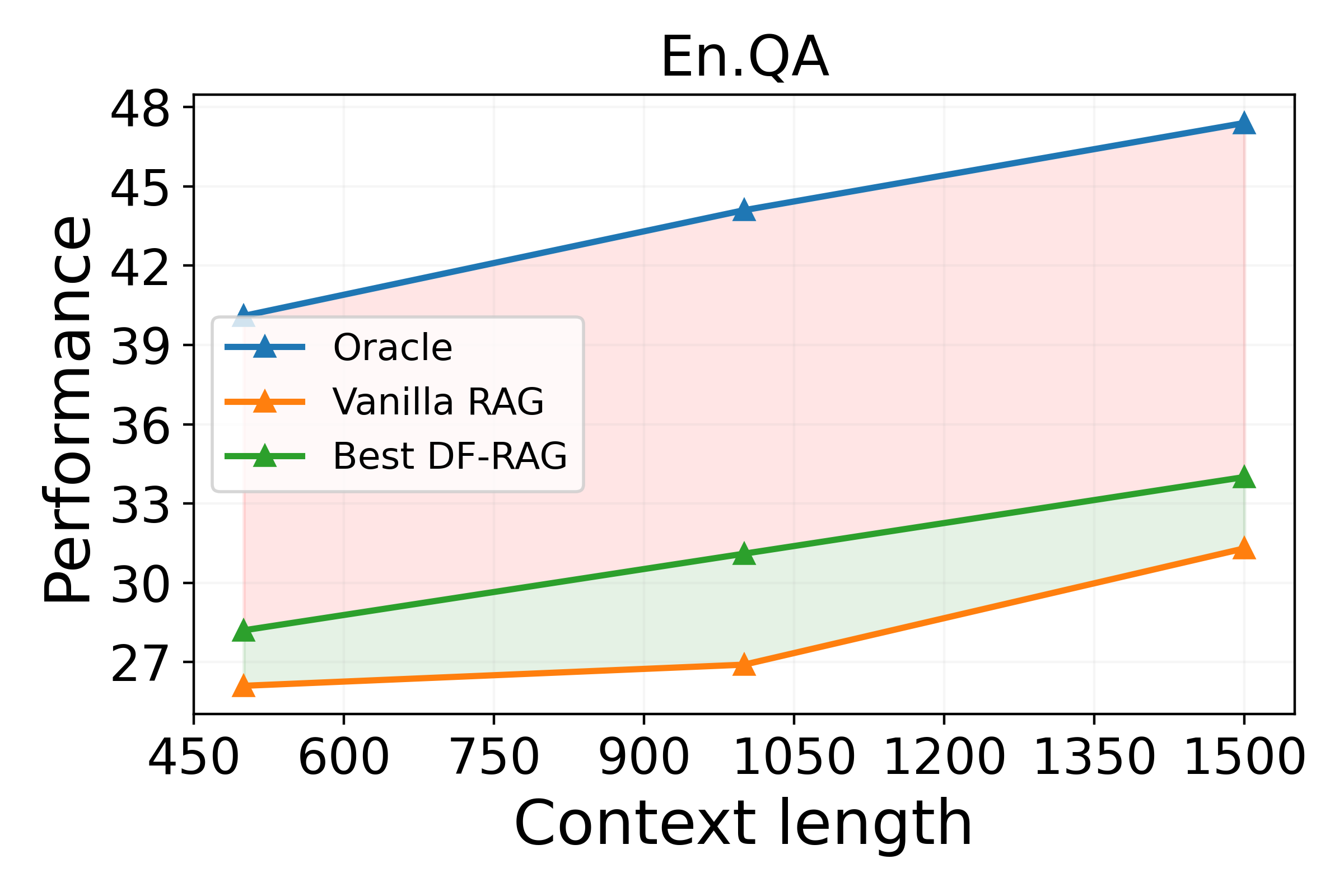}
  \end{subfigure}
  
  \vspace{0.1mm}
  
  \begin{subfigure}[t]{0.19\linewidth}
    \includegraphics[width=\linewidth]{ 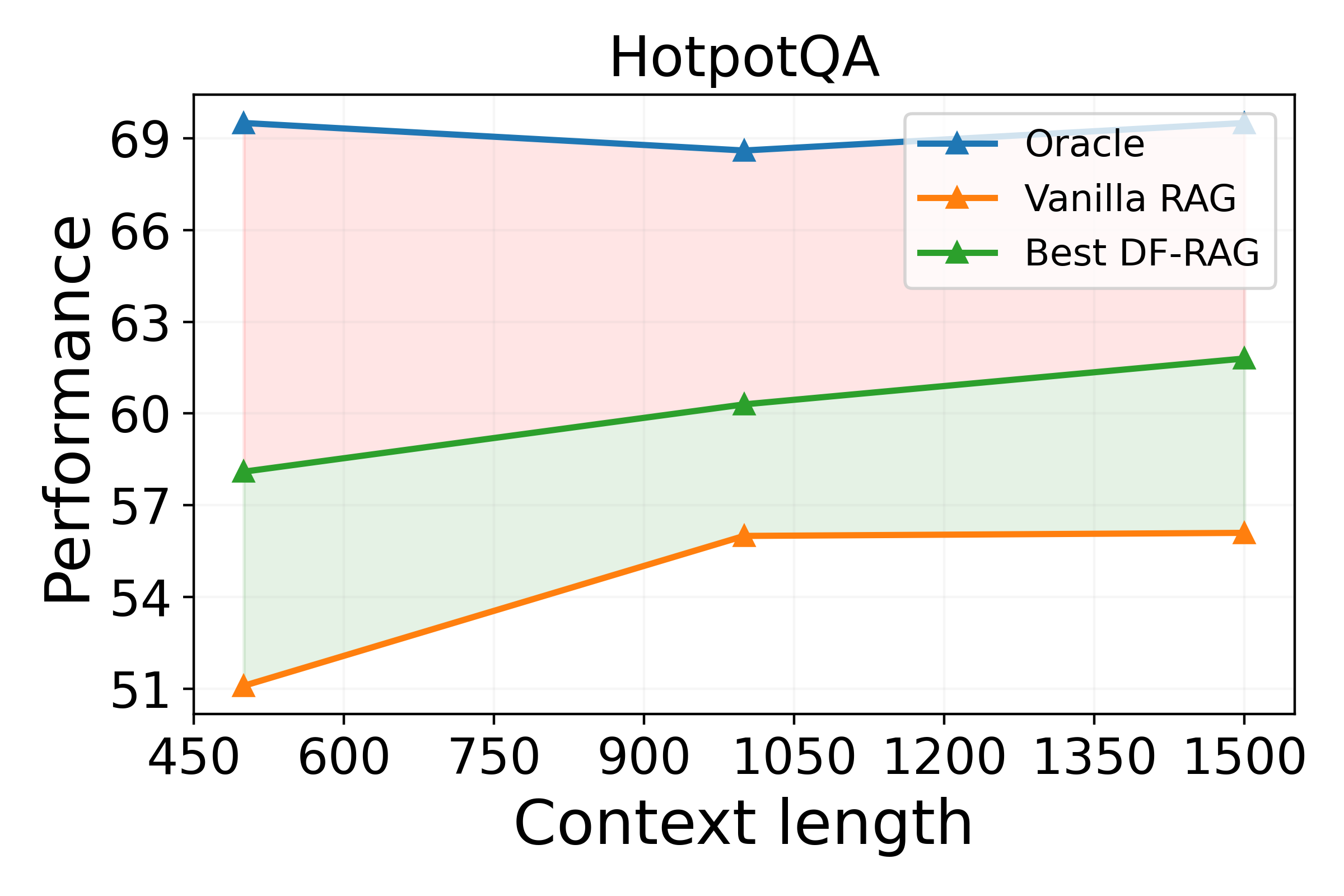}
  \end{subfigure}\hfill
  \begin{subfigure}[t]{0.19\linewidth}
    \includegraphics[width=\linewidth]{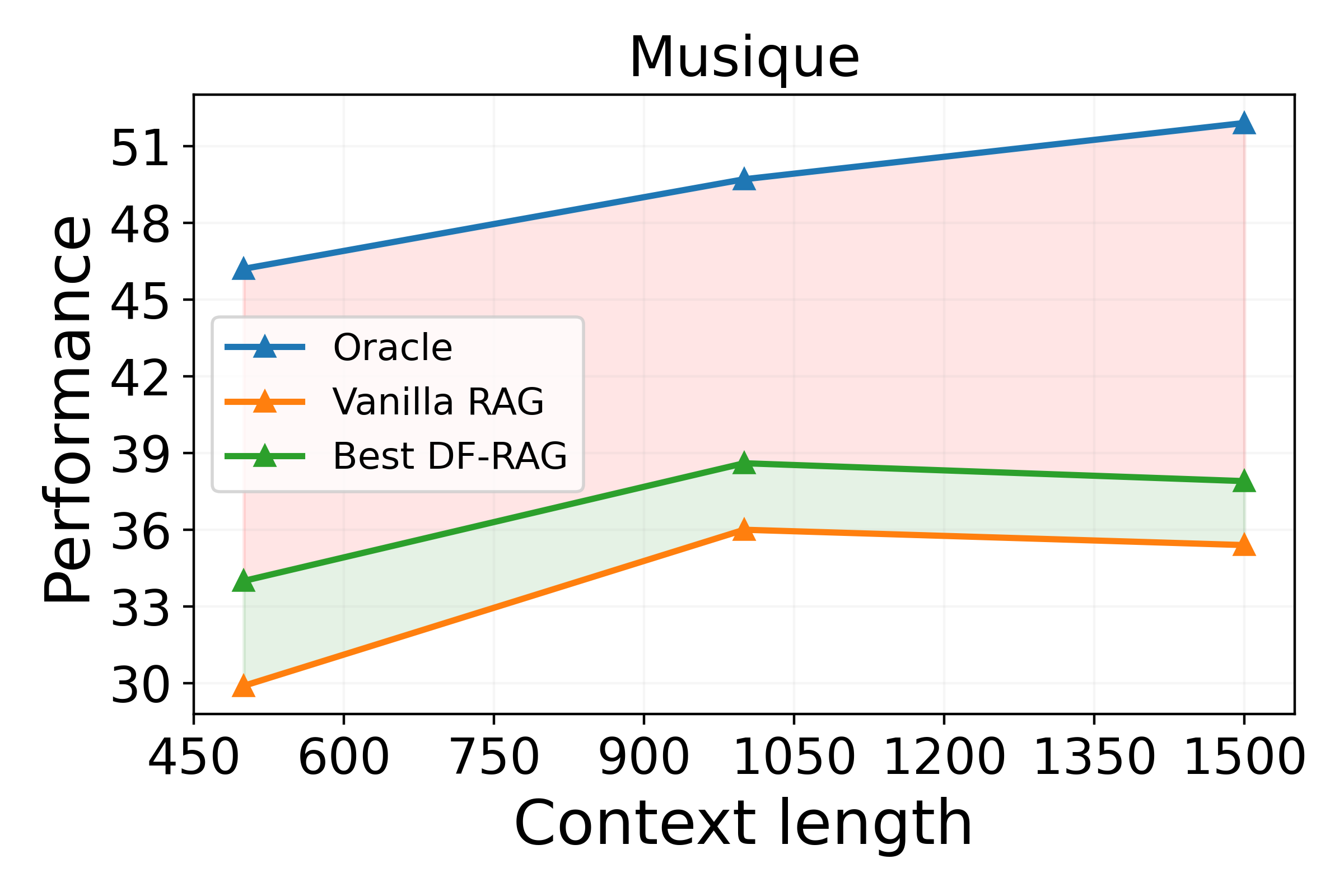}
  \end{subfigure}\hfill
  \begin{subfigure}[t]{0.19\linewidth}
    \includegraphics[width=\linewidth]{ 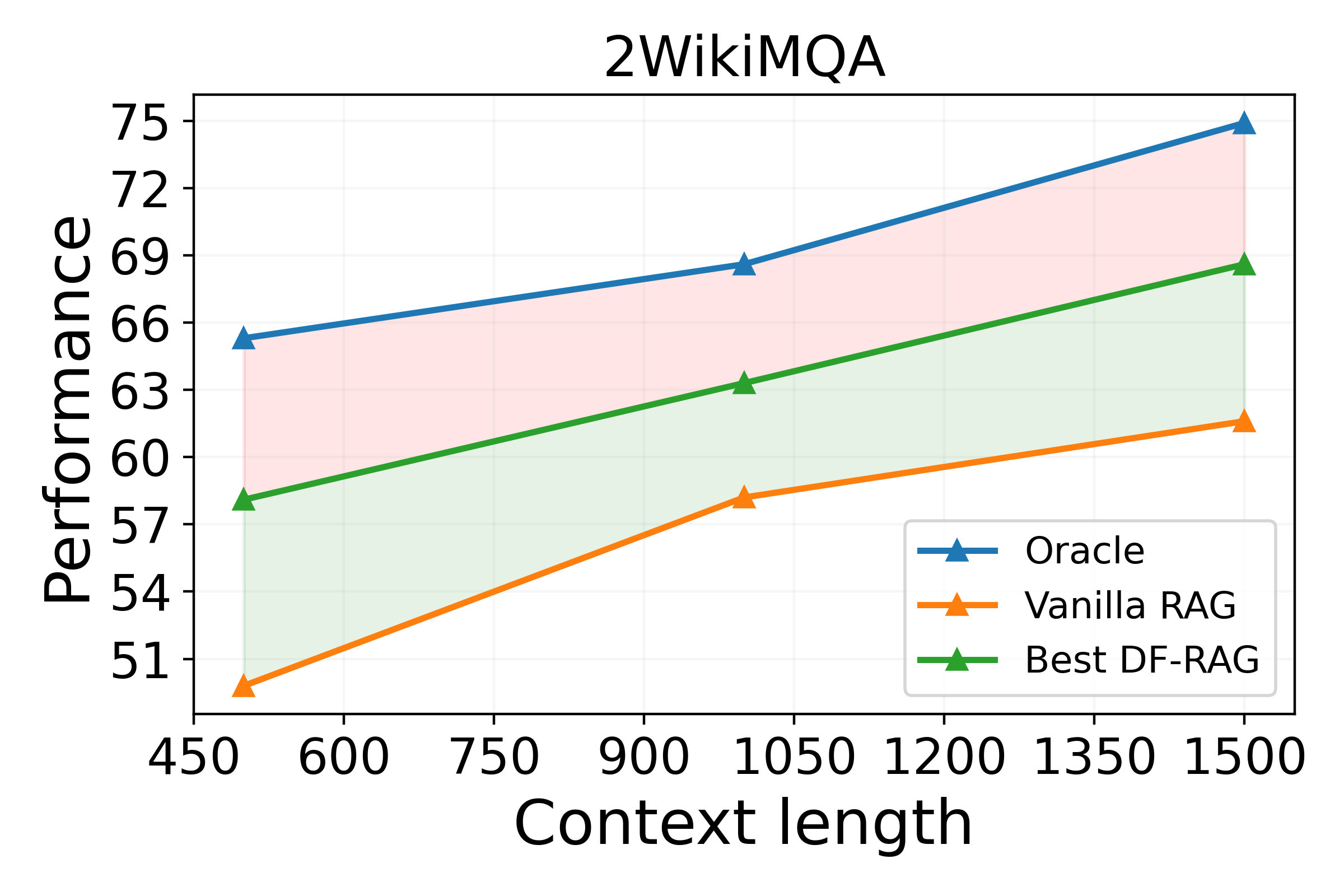}
  \end{subfigure}\hfill
  \begin{subfigure}[t]{0.19\linewidth}
    \includegraphics[width=\linewidth]{ 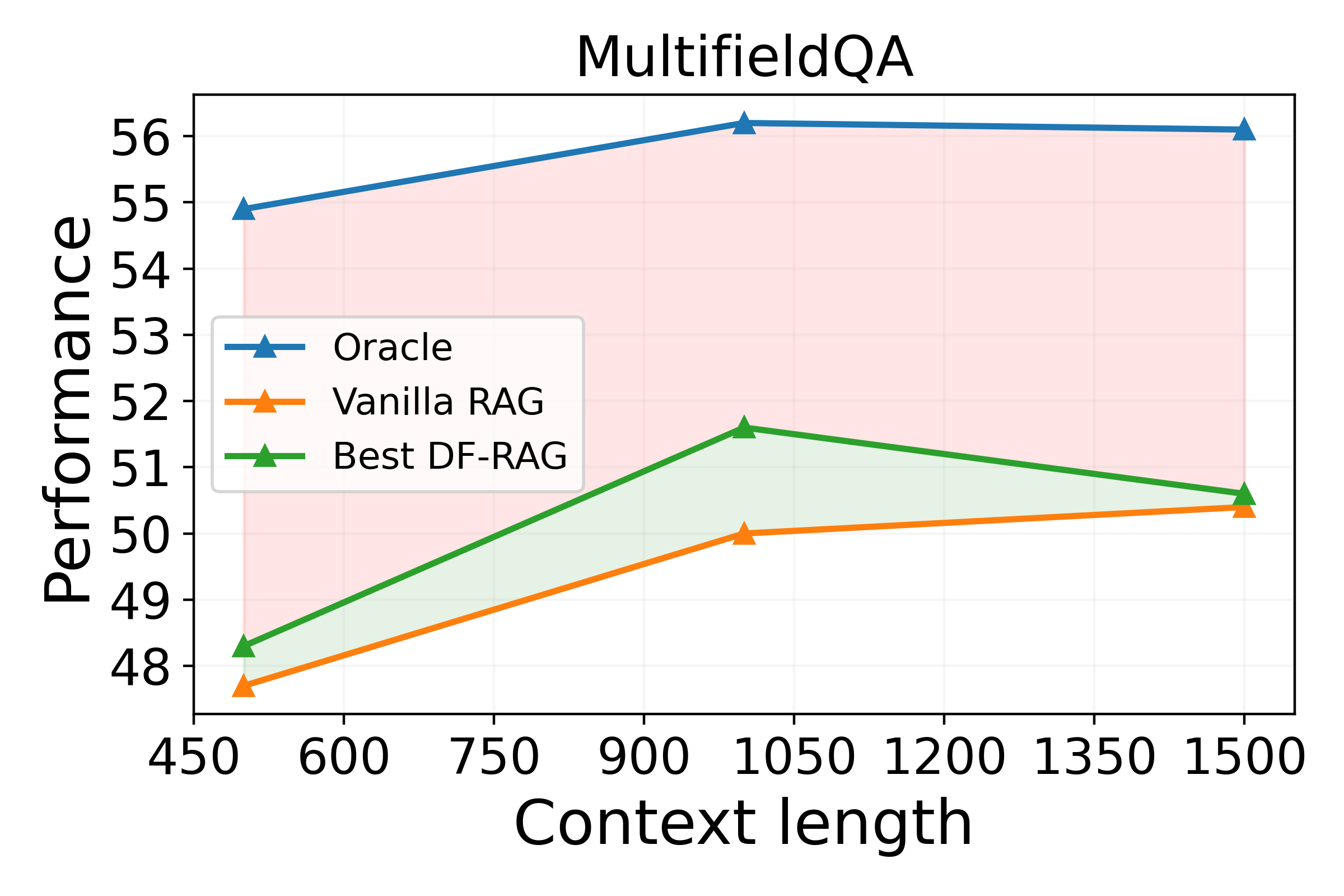}
  \end{subfigure}\hfill
  \begin{subfigure}[t]{0.19\linewidth}
    \includegraphics[width=\linewidth]{ 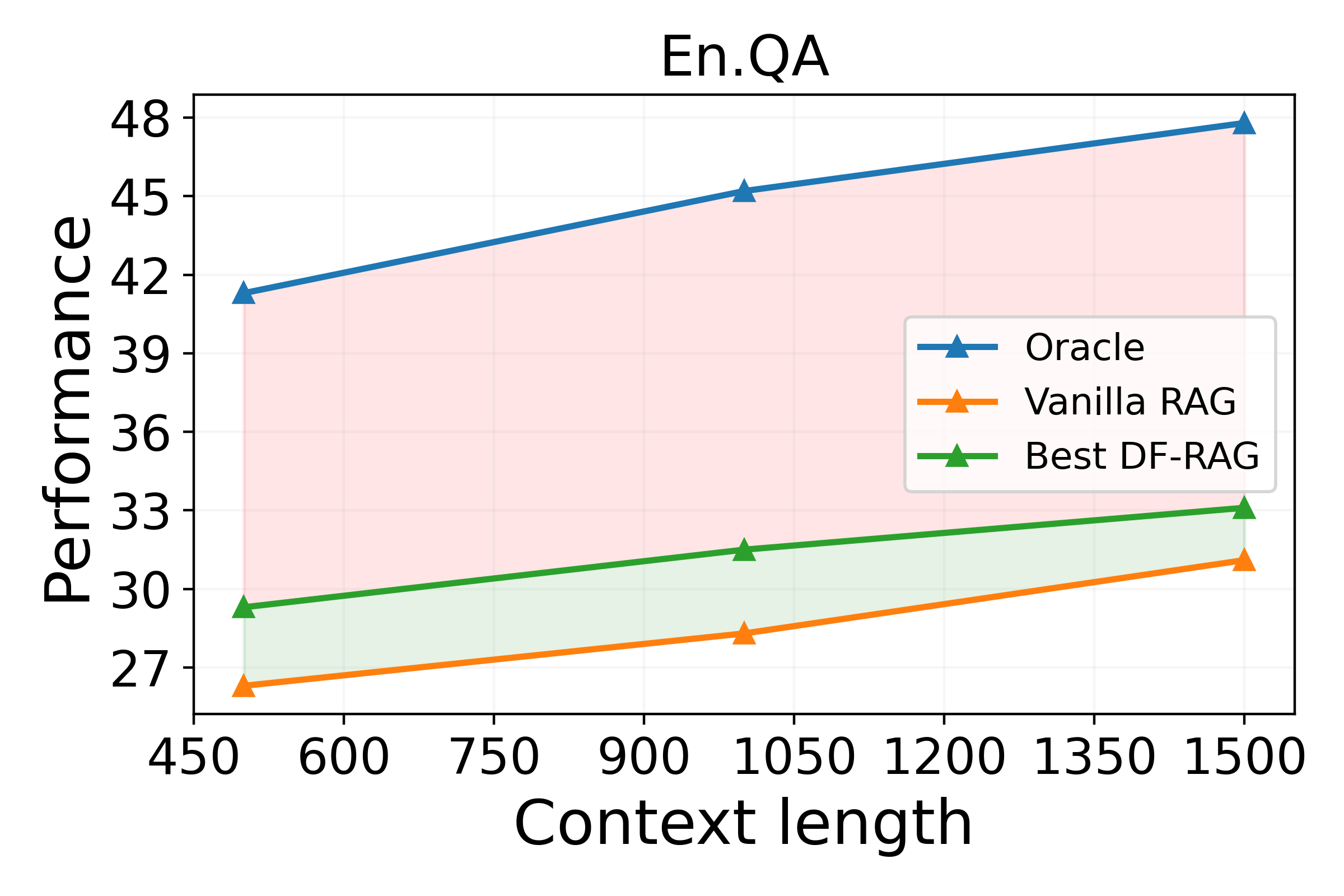}
  \end{subfigure}
  
  \caption{Performance ($F_1$ score) curves,  of Oracle, vanilla RAG, and best DF--RAG across context lengths on all five QA benchmarks. The first row shows Qwen 2.5 72B as the backbone LLM whereas the second row shows Llama 3.3 70B.}
  \label{fig:df-ragvsvanilla}
\end{figure*}

\subsection{DF-RAG}
\label{sec:dfr-res}
\noindent\textbf{DF-RAG beats RAG with $\lambda$-Fixed gMMR. } We observe that DF-RAG beats RAG with $\lambda$-Fixed gMMR consistently in various benchmarks across context lengths. Specifically, DF-RAG outperforms RAG with $\lambda$-Fixed gMMR with gains ranging from  $\sim2-10 \%$ across context lengths and models (see Figure  Table \ref{tab:performance_acl_dfrag_transposed}).
We generally observe larger gains with DF-RAG over RAG with $\lambda$-Fixed gMMR at longer context length settings (1500 words), suggesting that DF-RAG's ability of successfully picking the right $\lambda$ for each query helps longer context settings, as it gives more flexibility at retrieving information. However, we also observe that DF-RAG is outperformed by RAG with $\lambda$-Fixed gMMR in a small number of cases (MuSiQue and MultifieldQA datasets), but the magnitude of drop in performance is very minimal. All together, the performance shows DF-RAG's success and importance of its ability to accurately predict optimal $\lambda$ at test time.


\noindent\textbf{DF-RAG outperforms various RAG baselines.} 
Across benchmarks, \textsc{DF-RAG} consistently outperforms vanilla RAG, achieving up to +8.9\% absolute improvement at the 500-word context length, and up to +6.2\% and +9.5\% at the 1000- and 1500-word settings, respectively. \textsc{DF-RAG} maintains a clear advantage over vanilla RAG across all benchmarks, with more pronounced gains on HotpotQA and 2WikiMQA (see Figure~\ref{fig:df-ragvsvanilla}). Beyond vanilla RAG, \textsc{DF-RAG} also outperforms LongRAG and RAPTOR, two context length--independent baselines (LongRAG leverages global cues, while RAPTOR recursively summarizes the full context). Their results are shown in the last two rows of Table~\ref{tab:performance_acl_dfrag_transposed}. Notably, at the 1500-word context length, \textsc{DF-RAG} surpasses RAPTOR across all benchmarks and model families, and similarly outperforms LongRAG on every benchmark except MultifieldQA with Llama~3.3 70B and MuSiQue with Qwen~2.5 72B.

\noindent\textbf{DF-RAG approaches Oracle performance.} Beyond outperforming baselines across benchmarks, DF-RAG narrows the gap to the Oracle, revealing its potential across datasets, model families, and context lengths. To measure its effectiveness, we calculated the portion of the performance gap between vanilla RAG and the Oracle that DF-RAG closes. On multi-hop QA benchmarks, DF-RAG closes an average of 41.5\%, 32.1\%, and 47.7\% of this gap at 500, 1000, and 1500-word context lengths, respectively (see Figure \ref{fig:df-ragvsvanilla} and Table \ref{tab:performance_acl_dfrag_transposed}). For non–multi-hop QA benchmarks, DF-RAG covers on average $\sim$12-20\% of the gap. These results highlight two key insights: (1) DF-RAG effectively bridges reasoning gaps in multi-hop QA by adaptively balancing relevance and diversity through its retrieval component as it approaches Oracle performance (analyzed in Appendix \ref{sec:perf}). In one case, DF-RAG covers up to 91.3\% of the gap on a particular benchmark (2WikiMQA for Qwen 2.5 72B at 1500-word context length). (2) DF-RAG also maintains strong performance on complex non–multi-hop QA tasks, too (just below around 5\% of the Oracle for MultifieldQA). 
Together, the performance comparison of DF-RAG with the Oracle showcases DF-RAG's ability to adaptively predict optimal $\lambda$ at test time. Further analysis of DF-RAG's success and failure cases are presented in Appendix \ref{sec:ea}.  

\section{Discussion and Analysis}

\subsection{Sensitivity of $\lambda$ in selecting chunks}
To justify the difference in $\lambda$ sweeps in our experiments, we assess the sensitivity of $\lambda$ in selecting chunks within a given chunk-set. To illustrate this, we measure the Jaccard similarity index of the selected chunk-sets at each distinct $\lambda$, which provides an intuitive idea of the uniqueness of chunk-sets at neighboring $\lambda$ values. The Jaccard index (J) between two chunk sets $S_{\lambda_i}$ and $S_{\lambda_j}$ (sampled at $\lambda = i$ and $\lambda = j$) is measured as:
\[
\mathrm{J}(S_{\lambda_i}, S_{\lambda_j}) 
= \frac{\left| S_{\lambda_i} \cap S_{\lambda_j} \right|}
       {\left| S_{\lambda_i} \cup S_{\lambda_j} \right|}.
\]

\begin{figure*}[h]
  \centering
  \begin{subfigure}[t]{0.19\linewidth}
    \includegraphics[width=\linewidth]{ 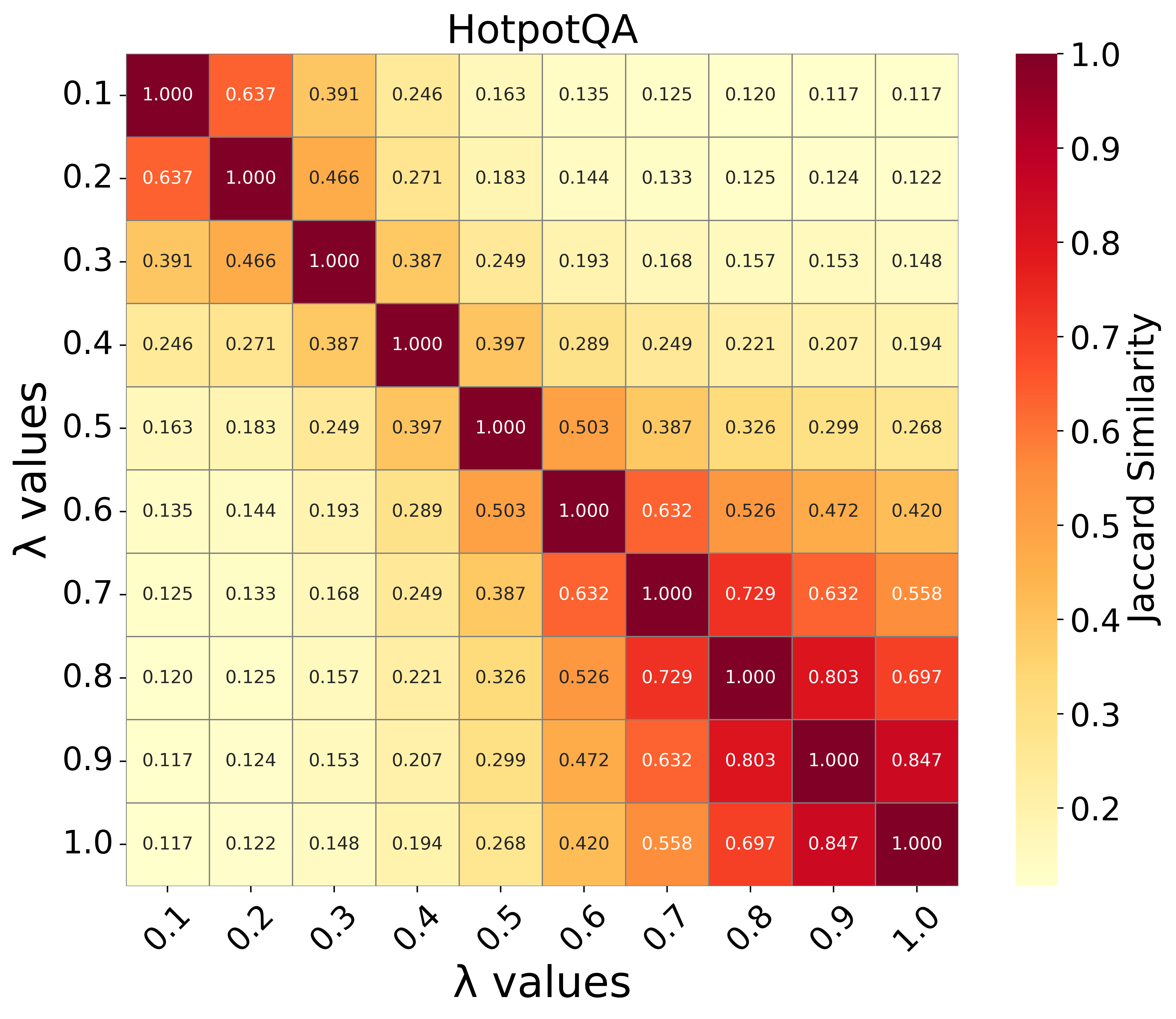}
  \end{subfigure}\hfill
  \begin{subfigure}[t]{0.19\linewidth}
   \includegraphics[width=\linewidth]{ 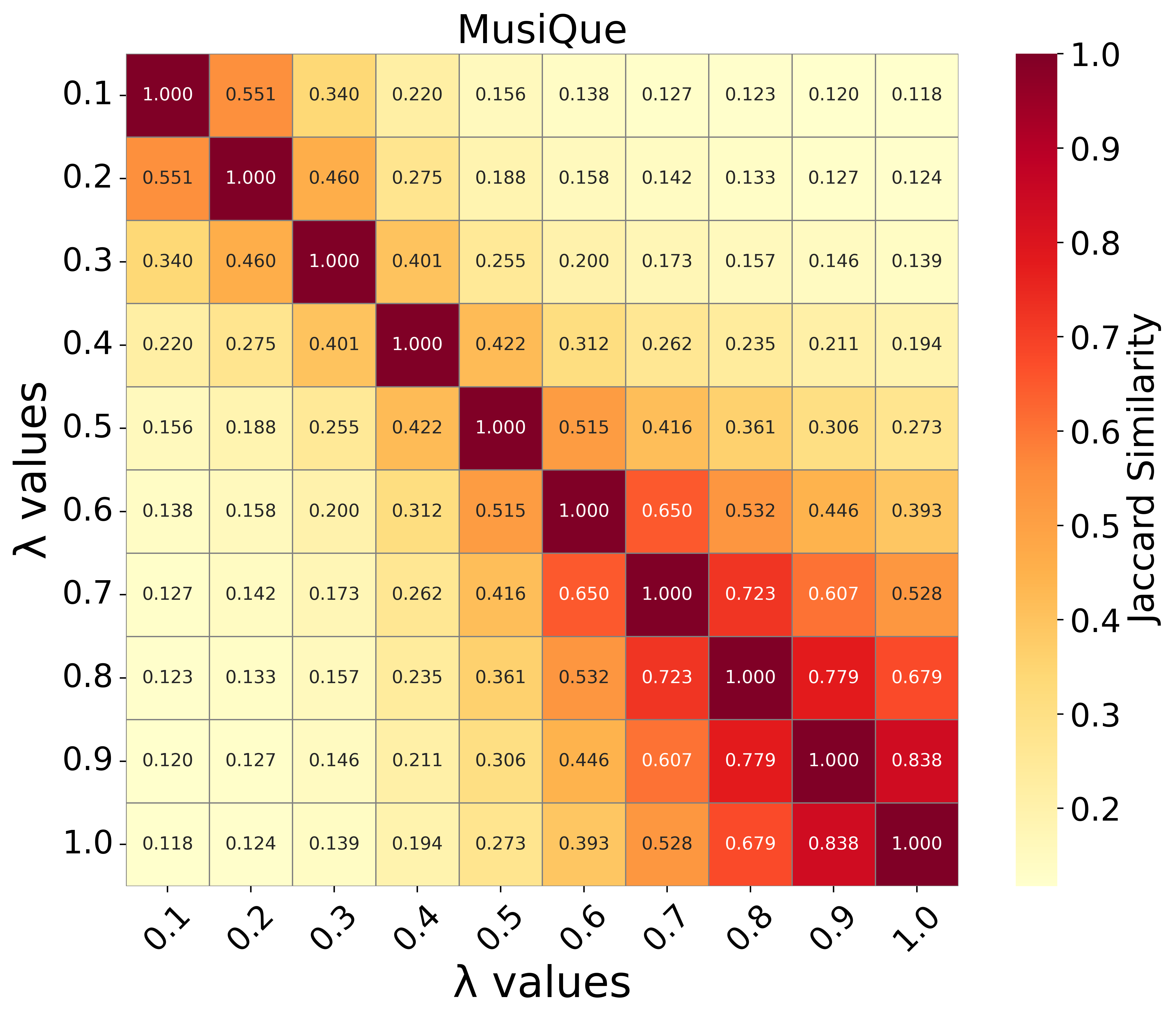}
  \end{subfigure}\hfill
  \begin{subfigure}[t]{0.19\linewidth}
    \includegraphics[width=\linewidth]{ 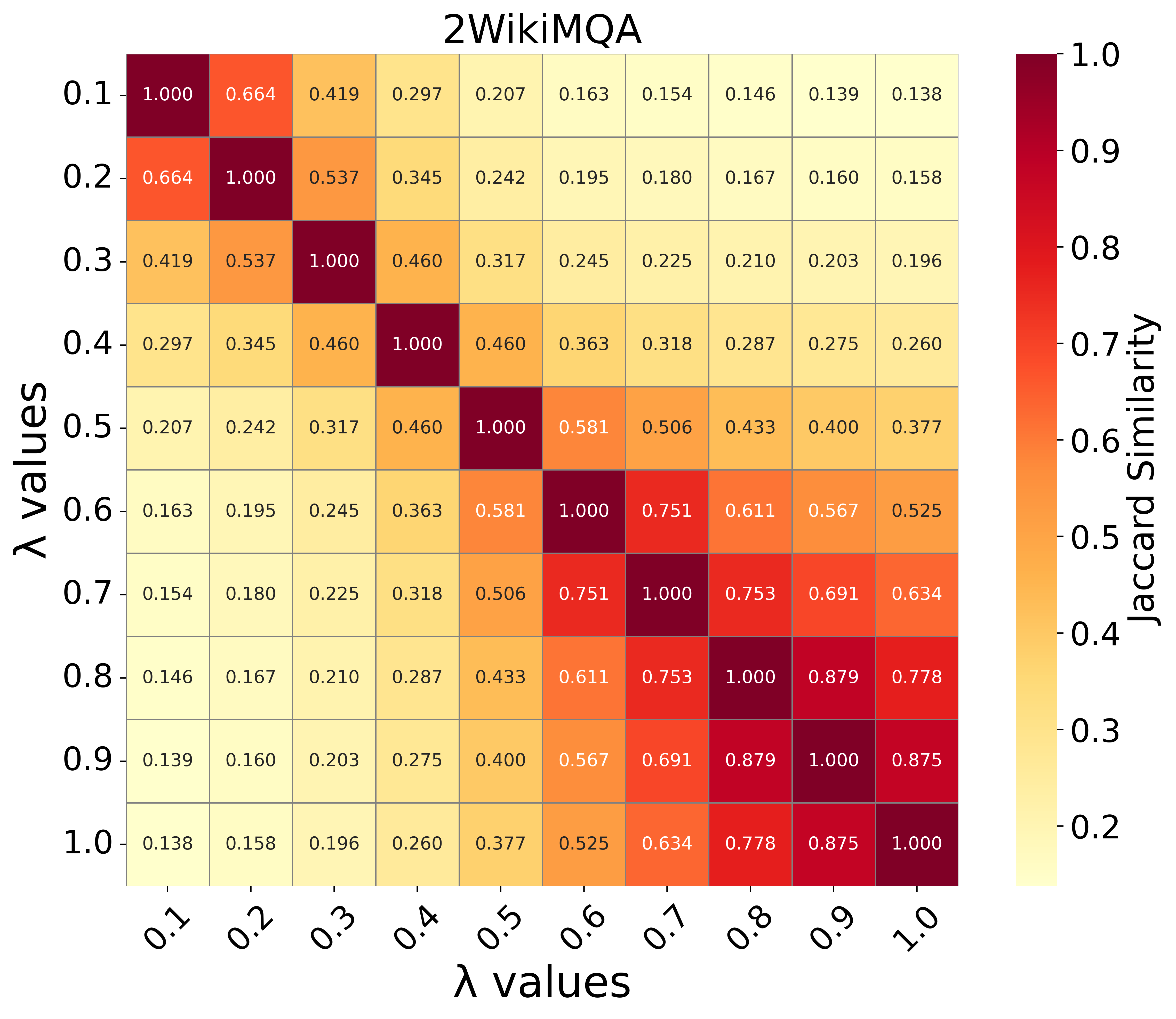}
  \end{subfigure}\hfill
  \begin{subfigure}[t]{0.19\linewidth}
    \includegraphics[width=\linewidth]{ 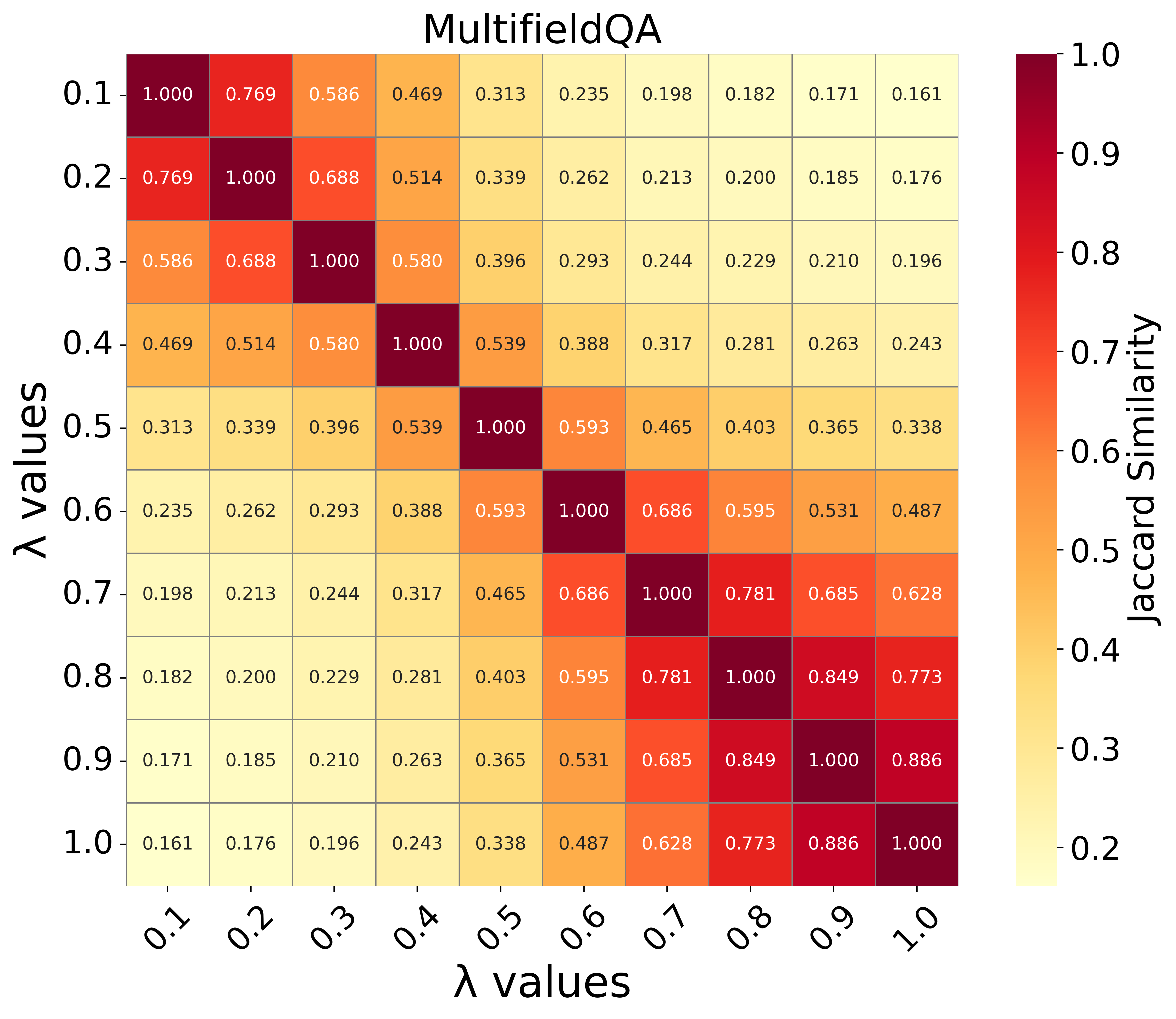}
  \end{subfigure}\hfill
  \begin{subfigure}[t]{0.19\linewidth}
    \includegraphics[width=\linewidth]{ 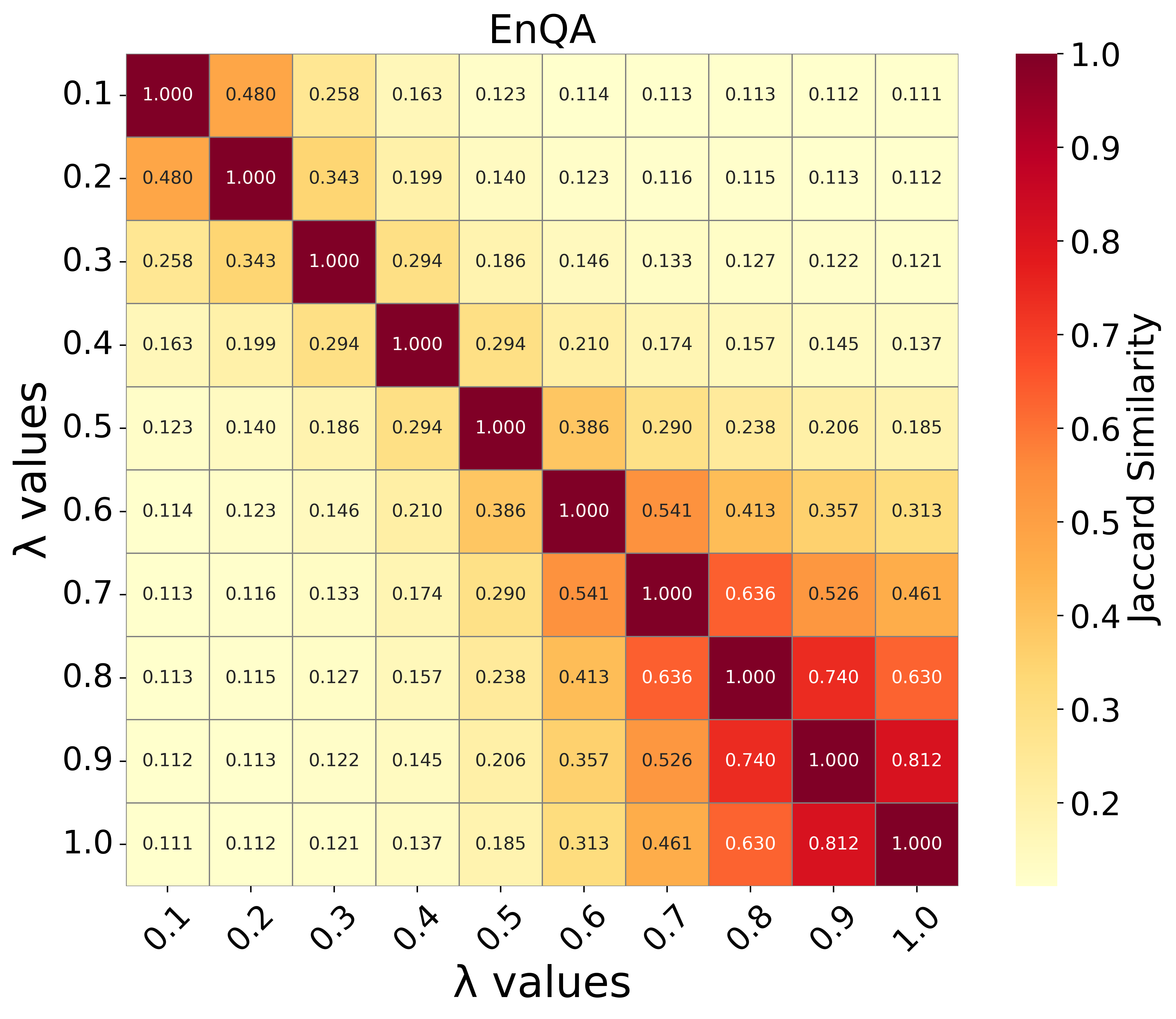}
  \end{subfigure}

  \caption{Heatmap showing Jaccard index values between candidate chunk-sets sampled at different $\lambda$. Darker color represents more common chunks between the corresponding chunk-sets.}
  \label{fig:lambda-sens}
\end{figure*}

Figure~\ref{fig:lambda-sens} shows the Jaccard index values between each pair of sampled chunk-sets. From the figure, we observe that even a small $\lambda$ increment of 0.1 introduces on average of about 40\% unique information.


\subsection{Latency Analysis}
We assess the latency of our DF-RAG by measuring its wall-clock runtime. For evaluation on latency, we use 20 randomly selected samples from MultifieldQA, which features reasonably long contexts, utilizing 300 word-chunks. While DF-RAG utilizes multiple LLM calls and its most significant time consuming component is the Evaluator which evaluates each chunk-set independently. However, due to the Evaluator's independent nature, its process can be parallelized which can significantly reduce the latency. The figure below shows how the runtime per sample varies with presence of parallelizable nodes. With more executable worker nodes, the Evaluator process can be parallelized, which reduces the latency.

\begin{figure}[H]
    \centering
    \includegraphics[width=0.9\linewidth]{ 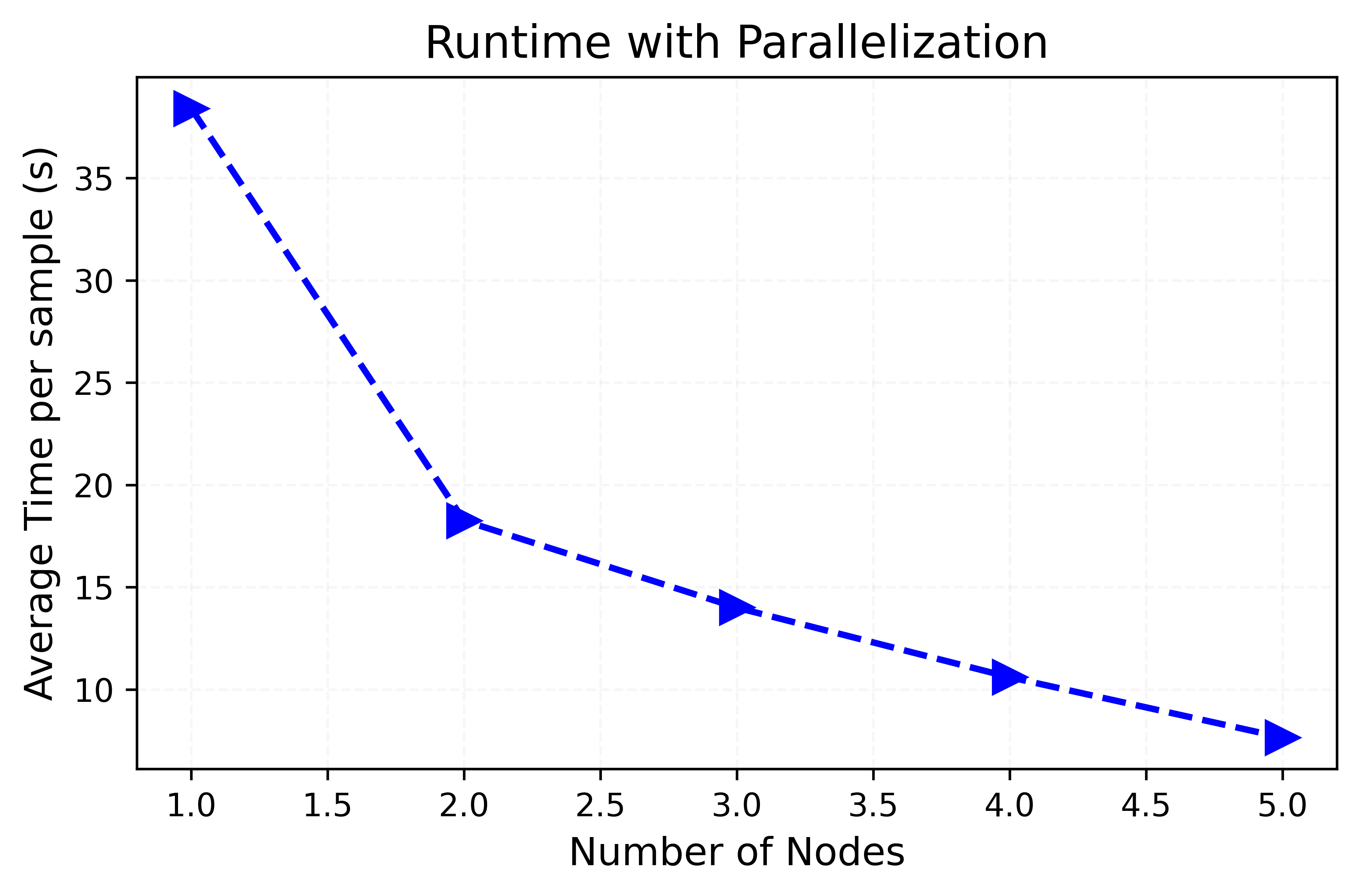}
    \caption{Runtime comparison of DF-RAG per sample with parallelizable worker nodes.}
    \label{fig:oracvslc}
\end{figure}

Furthermore, DF-RAG utilizes a mechanism that uses constant amount of token-processing with LLMs, regardless of the length of the context whereas, most advanced RAG systems' (as reported in our baselines) tokens process scale with the length of context supplied, due to extracting global cues, building graphs, etc. Therefore on very long contexts, DF-RAG becomes computationally efficient and a feasible approach, which is outlined by results obtained using our En.QA dataset, a dataset which HippoRAG and RAPTOR could not be evaluated on due to their demanding nature of time.

\subsection{Utilization of gMMR in retrieving from large corpus}

Although, like most advanced RAG systems, DF-RAG’s overall time complexity is primarily governed by the number of LLM calls required to generate an answer, retrieval complexity can become a bottleneck when operating over very large corpora or databases. In such settings, gMMR can be applied strategically as a re-ranking module on top of an initial Approximate Nearest Neighbor (ANN) retrieval stage. This design significantly reduces the computational overhead introduced by gMMR’s iterative selection process.

\subsection{Effectiveness of the Planner}
\label{sec:pln}
Since the Planner is responsible for breaking complex reasoning-intensive questions into intermediate steps, we evaluate its effectiveness based on the number of steps it generates per query. We hypothesize that an effective planner should avoid overplanning while breaking queries into a reasonable number of steps. Therefore, a good planner should produce steps close to the number of hops for multi-hop QA benchmarks. The Planner produces an average of \textbf{2.88} steps for 2WikiMQA, \textbf{2.71} for HotpotQA, and \textbf{2.81} for MuSiQue. These values closely align with the original number of hops in these datasets, supporting our claim that the Planner generates well-calibrated plans.

\subsection{Effectiveness of the Evaluator}
\label{sec:exc}

\begin{figure*}[h]
  \centering
  \begin{subfigure}[t]{0.3\linewidth}
    \includegraphics[width=\linewidth]{ 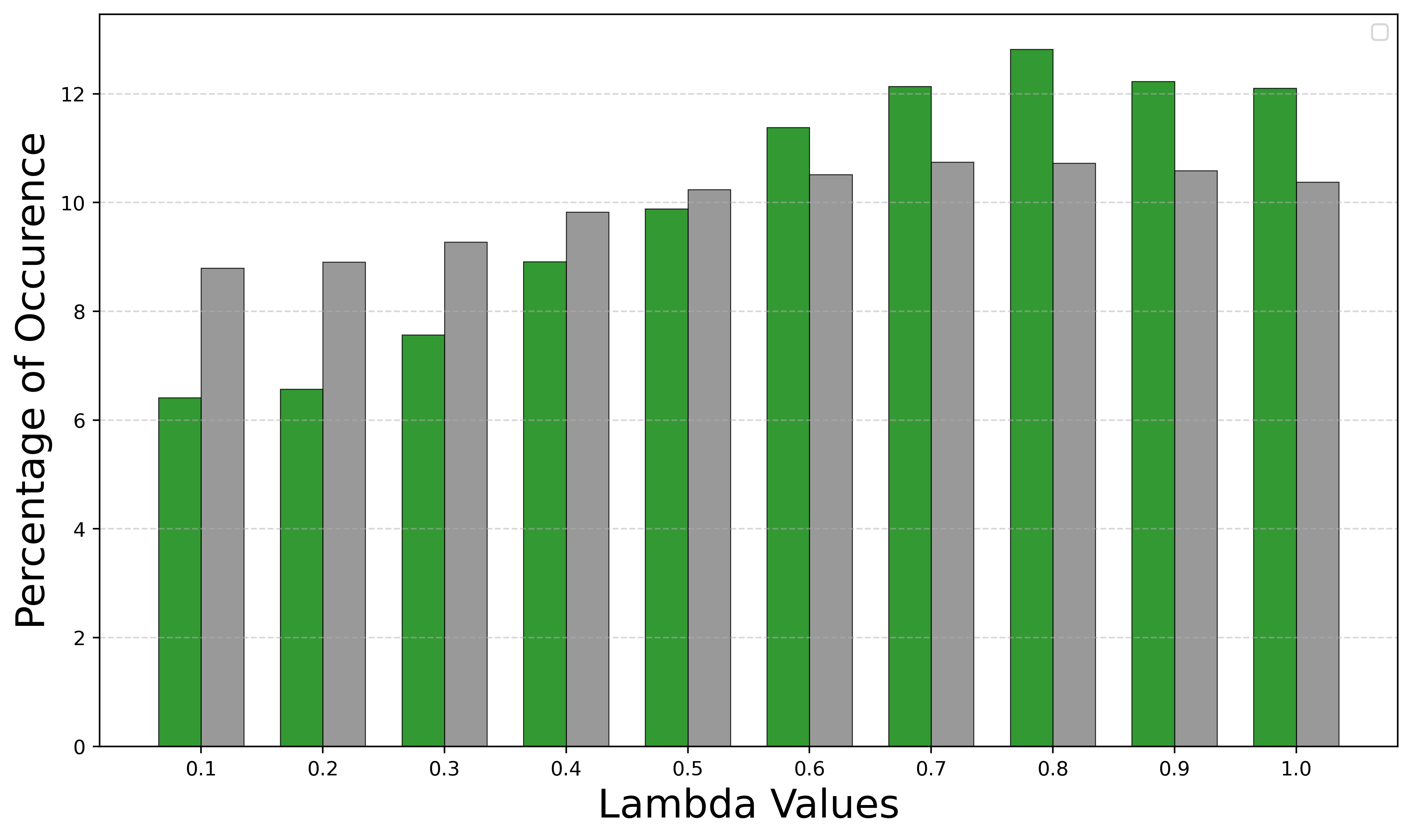}
    \caption{500 word context length}
  \end{subfigure}\hfill
  \begin{subfigure}[t]{0.3\linewidth}
    \includegraphics[width=\linewidth]{ 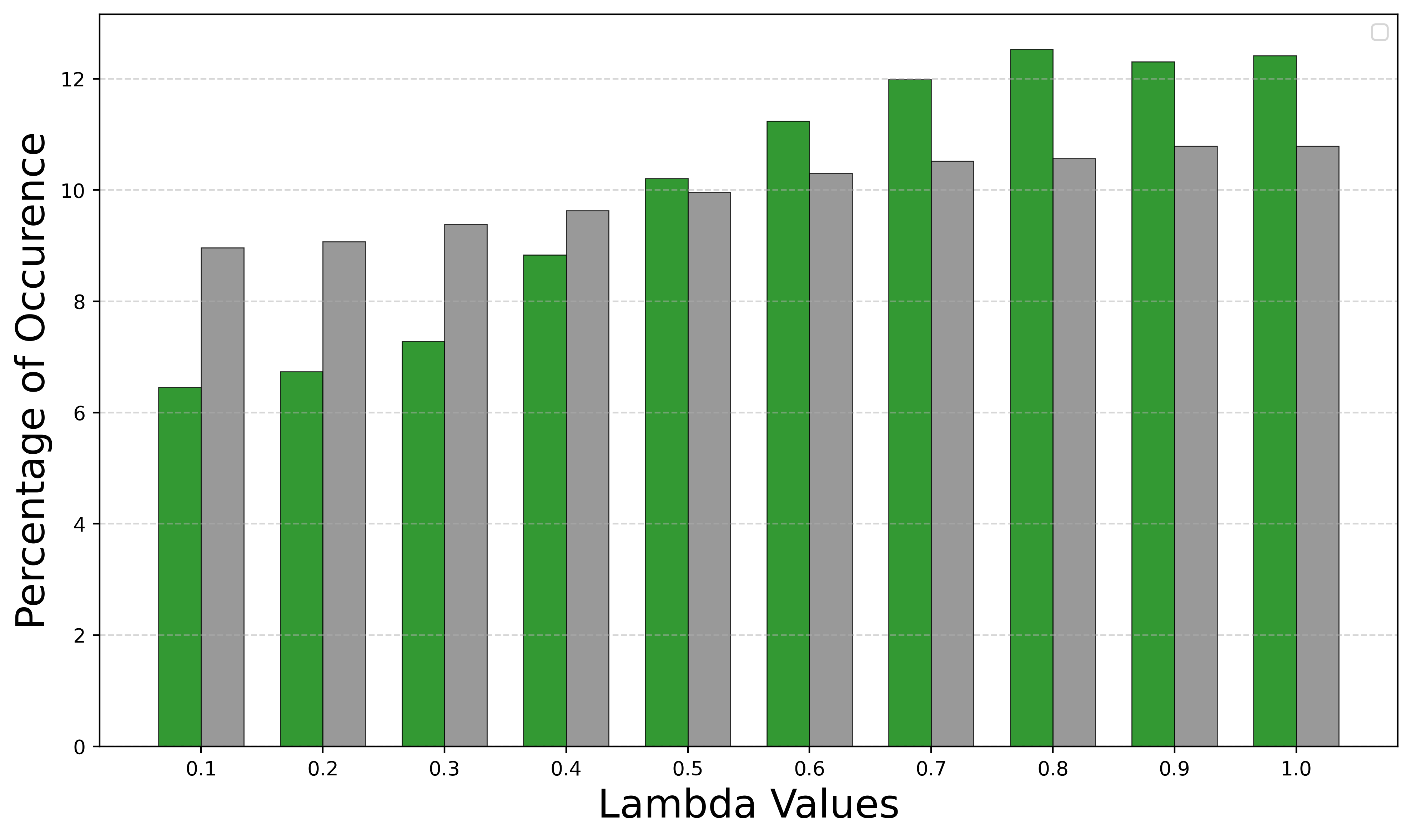}
    \caption{1000 word context length}
  \end{subfigure}\hfill
  \begin{subfigure}[t]{0.3\linewidth}
    \includegraphics[width=\linewidth]{ images/executor_val/dist_1000.png}
    \caption{1500 word context length}
  \end{subfigure}
  
  \vspace{0.1mm}
  
  \caption{Distribution of $\lambda$ values from the Evaluator of DF-RAG (green) and the Oracle (gray). Both exhibit smooth distributions, though the Oracle’s is slightly smoother, reflecting the source of DF-RAG’s performance gap.}
  \label{fig:two-by-three}
\end{figure*}
We evaluate the effectiveness of the Evaluator in our DF-RAG pipeline by examining the distribution of its predicted $\lambda$ values from the Evaluator across datasets. An effective Evaluator should assign a diverse range of $\lambda$ values, reflecting its ability to adapt retrieval diversity to the unique requirements of each query. Such behavior would generate a smooth and broad distribution of $\lambda$s. In contrast, a poorly performing Evaluator would produce a narrow, concentrated distribution -- indicating that it defaults to similar $\lambda$ values for most queries, failing to capture query-specific diversity needs and effectively outputting non-informative predictions. Figure \ref{fig:two-by-three} shows the distribution of $\lambda$s from Evaluator and from our Oracle.

\subsection{Effectiveness of appending the Incremental Context} 
\label{sec:ic}
Our results show that the DF-RAG variant outperforms the DF-RAG + IC variant in several settings -- for example, on HotpotQA, 2WikiMQA and MultifieldQA and En.QA at the 500-word context length with Llama 3.3 70B; on HotpotQA, MultifieldQA and En.QA at 1500 words with Qwen 2.5 72B. The IC augmentation was intended to provide a concise rationale, which—following findings from Chain-of-Thought (CoT) \cite{10.5555/3600270.3602070}—should have enhanced performance. However, our experiments reveal that in some cases (notably always on MultifieldQA and En.QA), appending the IC leads to slightly lower performance. We attribute this finding to two factors: 1) For non-multi-hop reasoning, extended reasoning chains are not only computationally inefficient \cite{chen2024not} but also potentially detrimental to performance \cite{hassid2025don}. Drawing a parallel to this ``overthinking'' problem, the IC could make less complicated queries seemingly complicated through excessive planning steps - deviating LLMs reasoning line of bridging information that hurts performance. Consequently, mitigation strategies developed to curb model overthinking may also be applicable to DF-RAG \cite{fang2025thinkless}. 2) Our experiments use small contexts (500-1500 words) where the information required to reach a multi-hop answer is already highly condensed, leaving little room for condensed reasoning to provide additional benefit.

\section{Conclusion}

In this work, we addressed the challenges RAG pipelines face on complex, reasoning-intensive QA
tasks, identifying their reliance on cosine similarity as a key limitation that leads to redundant contexts.  To overcome this, we introduced a retrieval function that balances diversity and relevance by a control parameter -- $\lambda$. We incorporated this retrieval
function at a dataset-level which showed performance improvements over vanilla RAG. With our given retrieval mechanism, we then designed an Oracle to understand the upper bound of performance improvement possible by controlling our $\lambda$ at a query level, which revealed up to 18\% absolute performance gains from vanilla RAG. Building on these insights, we proposed DF-RAG, which employs query-aware diversity by predicting the optimal $\lambda$ for each query at test time without prior task-specific tuning. To test our system's performance on complex reasoning-intensive QA tasks, we evaluated DF-RAG on multi-hop QA and long-context non–multi-hop QA datasets across various context lengths and 2 model families. DF-RAG routinely outperforms strong baselines across context lengths and recovers up to 91.3\% of the performance gap between vanilla RAG and the Oracle. Furthermore, DF-RAG uses constant token processing and can achieve faster execution through parallelization. Our results highlight the importance of diversity-focused retrieval for improving RAG pipelines on QA tasks and direct future directions for further innovation at closing the gap to Oracle performance.

\section*{Limitations}
While DF-RAG brings about performance improvements, there are limitations in our approach that merit discussion. 
First, given the large number of experiments conducted (over 180 in total), we only benchmarked using two LLM families (Qwen and Llama). Second, while we benchmarked on 1,101 samples across five datasets from LongBench and $\infty$Bench, leveraging a wider range of datasets with more samples from such unified suites would enable stronger assessments of generalizability. Third, we deliberately focused on fine-tuning-free baselines while benchmarking DF-RAG's performance for fair comparison; investigating its performance and benefits over training-dependent methods such as Self-RAG \cite{asai2024selfrag} remains as a future work. Finally, a central component of DF-RAG is its Planner and Evaluator, which rely on LLMs. While we conduct systematic ablation studies on both these components to evaluate their performances, improved prompting and other test time strategies \cite{snell2024scaling} may improve the effectiveness and performance of DF-RAG.

\section*{Ethical Considerations}

Our research underwent a standard internal review. We used publicly available datasets and LLMs in accordance with their intended scope, and we refer readers to the original dataset papers for details on anonymization, preprocessing, and annotator information. All artifacts (existing baselines and benchmarks) used in this this paper are available under their respective open-source licenses (except for LongRAG; we only reproduce their code). Our experiments focus exclusively on the English language; however, the proposed methodology makes no assumptions about the language of operation and we encourage future work to explore the applicability of our approach in multilingual settings. Furthermore, we acknowledge the propagation of various biases through LLM-based systems and while our methods do not study the implications of biases in our system, we recommend that any system implementing our methodology undergo rigorous testing of (1) the documents retrieved to prevent the propagation of biases, and (2) the models used to generate answers to ensure they do not promote biases during generation. Such systems should also be subject to continuous monitoring. We also acknowledge the use of ChatGPT for fixing grammatical and spelling errors.

\newpage

\bibliography{custom}

\newpage

\appendix
\onecolumn
\section{Methods}

\label{sec:appendix}

\subsection{Dataset Statistics}
\label{dataset-stat}

The dataset statistics are presented in Table \ref{tab:dataset_stats}.
\begin{table}[H]
\centering
\begin{tabular}{lcc}
\toprule
\textbf{Dataset} & \textbf{No. of Instances} & \textbf{Avg. Length} \\
\midrule

\multicolumn{3}{c}{\textit{LongBench} \cite{bai-etal-2024-longbench}} \\
\midrule
HotpotQA         & 200                    & 9,151                  \\
2WikiMQA      & 200                    & 4,887                   \\
MuSiQue          & 200                      & 11,214                   \\
MultifieldQA      & 150                      & 4,559                    \\

\midrule
\multicolumn{3}{c}{\textit{$\infty$Bench} \cite{zhang-etal-2024-bench}} \\
\midrule
EN.QA            & 351 & 192,600              \\
\bottomrule
\end{tabular}
\caption{Total number of queries and average length of contexts in number of words for each QA dataset.}
\label{tab:dataset_stats}
\end{table}

\subsection{Design choice of selecting chunks}
Despite the many possible strategies for constructing a chunk set, we adopt our proposed approach of selecting a fixed-size set of \( n = 5 \) chunks from the document chunk collection \( \mathcal{C} \) at a single \( \lambda \) value. This design enables a one-time computation of gMMR scores over all candidate chunks in \( \mathcal{C} \) at a particular $\lambda$, followed by a single evaluation of the resulting chunk-set using the Evaluator. However, it is also possible that the chunk-set selection strategy can be more adaptive at the cost of more computation, like iteratively selecting the best chunk and incrementally building the set which would requiring up to \( n \) additional Evaluator calls per query.

\subsection{System Prompts and Reproducibility}
\label{sec:systemprompts}
We release all our system prompts starting from the Planner, Evaluator and the Generator. Figure \ref{fig:planner} shows the prompt used for the Planner in DF-RAG. Figure \ref{fig:exec} shows the prompt used by our Evaluator. The Evaluator utilizes few shot prompting and the few shot examples specifically utilized are further detailed in Figure \ref{fig:few-shot} and Figure \ref{fig:few-shot-contd}. Our system can be reproduced by following these system prompts and the architecture proposed in Figure \ref{fig:df-ragvsvanilla}. We use the vanilla Hugging Face Instruct versions for all backbone LLMs.

\section{Additional Experimental Results}

\subsection{RAG with $\lambda$-Fixed gMMR:}
\label{sec:RAGWMMRcontd}
\begin{figure*}[h]
    \centering

    \parbox[c][1em][c]{2em}{}
    \begin{minipage}[c][1em][c]{\textwidth}
        \centering
        \begin{subfigure}[b]{0.3\textwidth}
            \centering\textbf{context length = 500 words}
        \end{subfigure}%
        \hfill%
        \begin{subfigure}[b]{0.3\textwidth}
            \centering\textbf{context length = 1000 words}
        \end{subfigure}%
        \hfill%
        \begin{subfigure}[b]{0.3\textwidth}
            \centering\textbf{context length = 1500 words}
        \end{subfigure}
    \end{minipage}
    \vspace{0.2cm}

    \parbox[c][4cm][c]{2em}{\rotatebox{90}{\centering \textbf{Hotpot QA}}}%
    \begin{minipage}[c][4cm][c]{0.9\textwidth}
        \centering
        \begin{subfigure}[b]{0.33\textwidth}
            \centering
            \includegraphics[width=\textwidth]{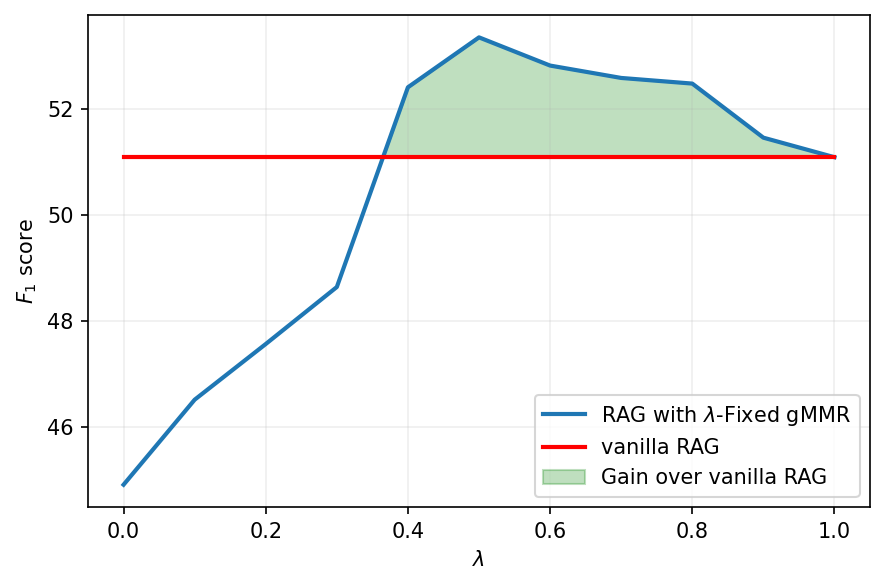}
            \label{fig:plot1}
        \end{subfigure}%
        \hfill%
        \begin{subfigure}[b]{0.33\textwidth}
            \centering
            \includegraphics[width=\textwidth]{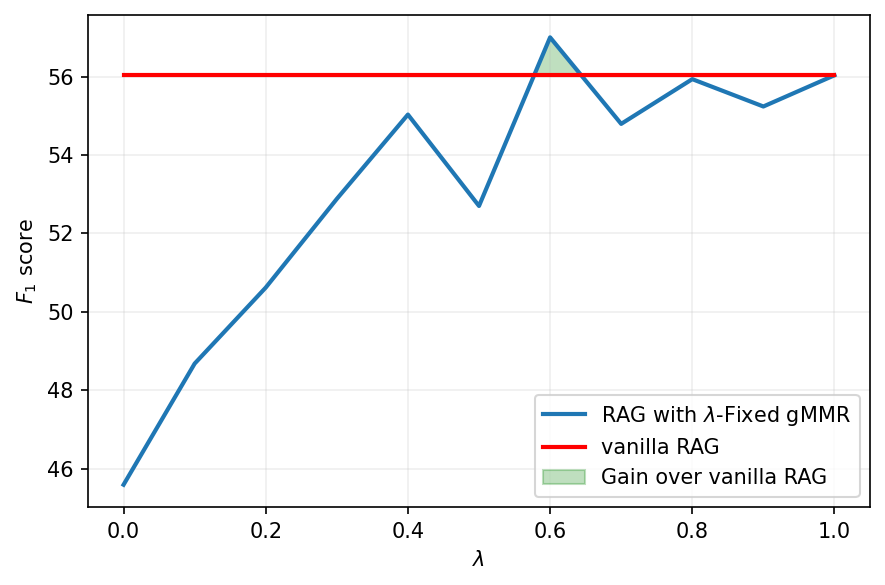}
            \label{fig:plot2}
        \end{subfigure}%
        \hfill%
        \begin{subfigure}[b]{0.33\textwidth}
            \centering
            \includegraphics[width=\textwidth]{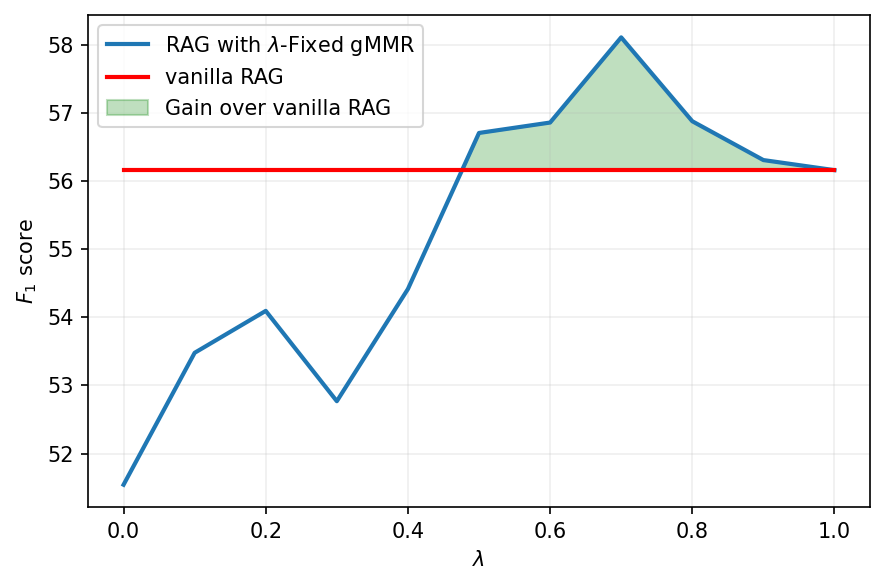}
            \label{fig:plot3}
        \end{subfigure}
    \end{minipage}

    \vspace{0.5cm}

    \parbox[c][4cm][c]{2em}{\rotatebox{90}{\centering \textbf{MuSiQue}}}%
    \begin{minipage}[c][4cm][c]{0.9\textwidth}
        \centering
        \begin{subfigure}[b]{0.33\textwidth}
            \centering
            \includegraphics[width=\textwidth]{ 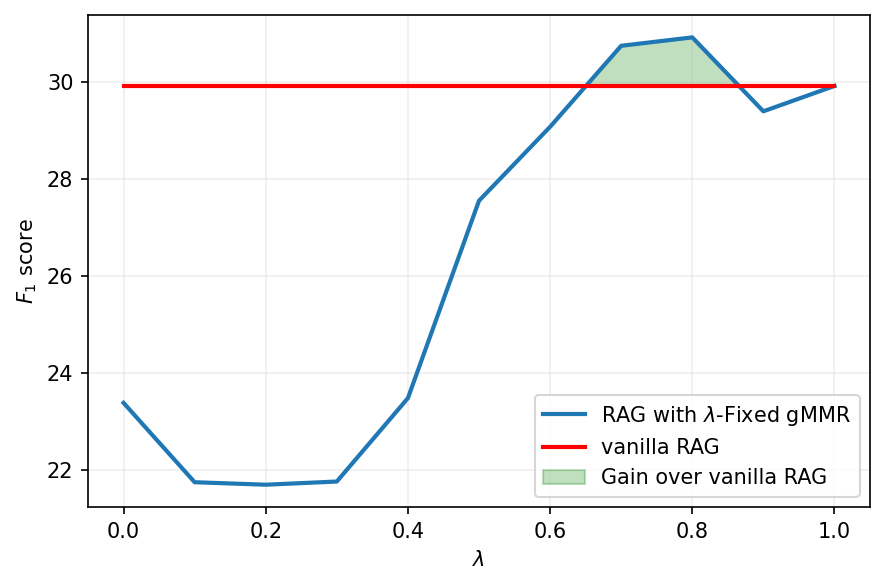}
            \label{fig:plot4}
        \end{subfigure}%
        \hfill%
        \begin{subfigure}[b]{0.33\textwidth}
            \centering
            \includegraphics[width=\textwidth]{ 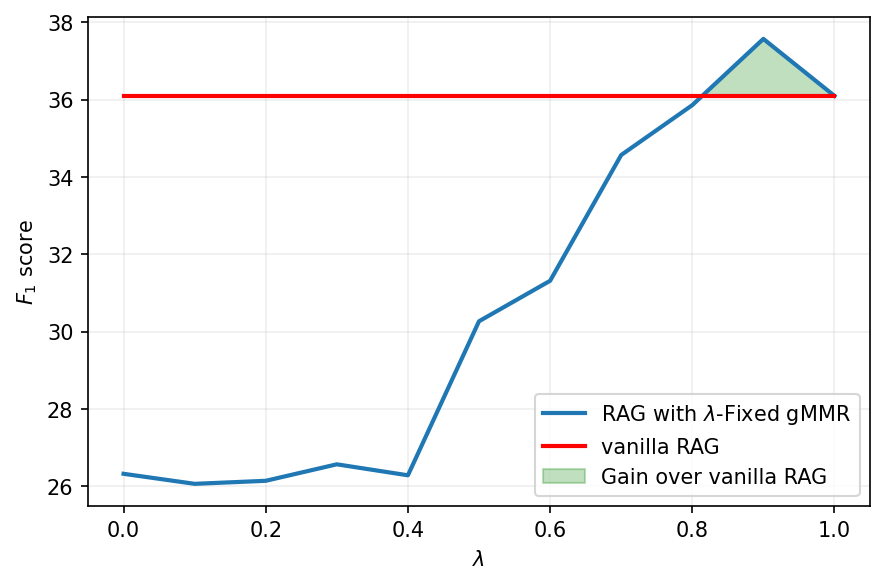}
            \label{fig:plot5}
        \end{subfigure}%
        \hfill%
        \begin{subfigure}[b]{0.33\textwidth}
            \centering
            \includegraphics[width=\textwidth]{ 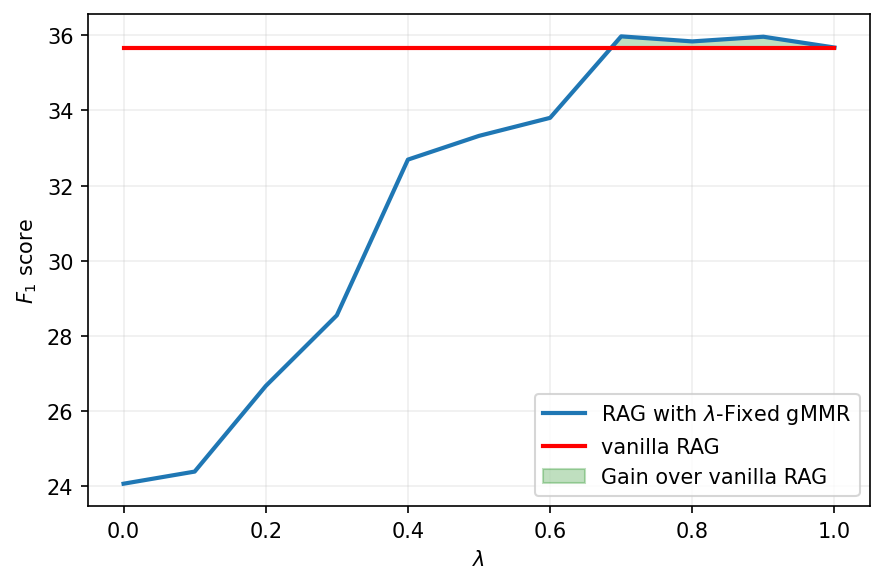}
            \label{fig:plot6}
        \end{subfigure}
    \end{minipage}

    \vspace{0.5cm}

    \parbox[c][4cm][c]{2em}{\rotatebox{90}{\centering \textbf{2WikiMQA}}}%
    \begin{minipage}[c][4cm][c]{0.9\textwidth}
        \centering
        \begin{subfigure}[b]{0.33\textwidth}
            \centering
            \includegraphics[width=\textwidth]{ 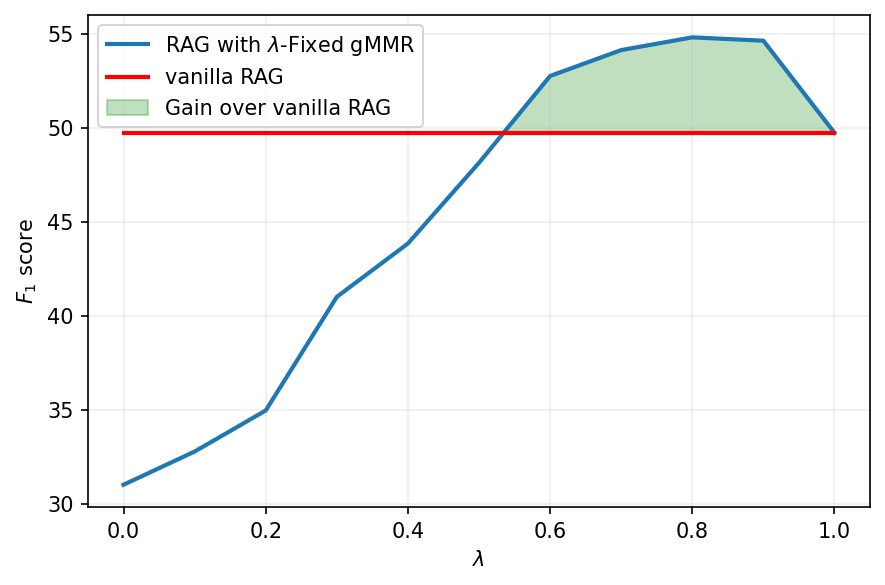}
            \label{fig:plot10}
        \end{subfigure}%
        \hfill%
        \begin{subfigure}[b]{0.33\textwidth}
            \centering
            \includegraphics[width=\textwidth]{ 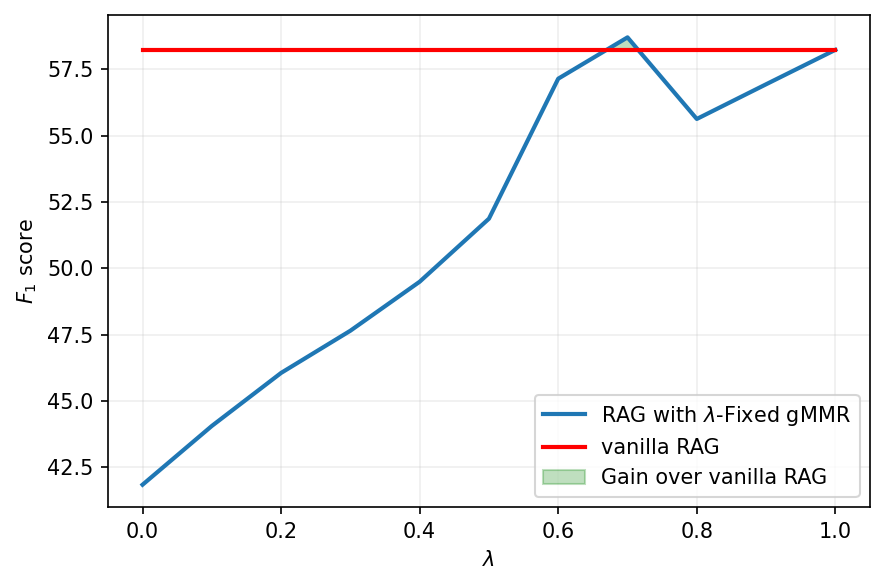}
            \label{fig:plot11}
        \end{subfigure}%
        \hfill%
        \begin{subfigure}[b]{0.33\textwidth}
            \centering
            \includegraphics[width=\textwidth]{ 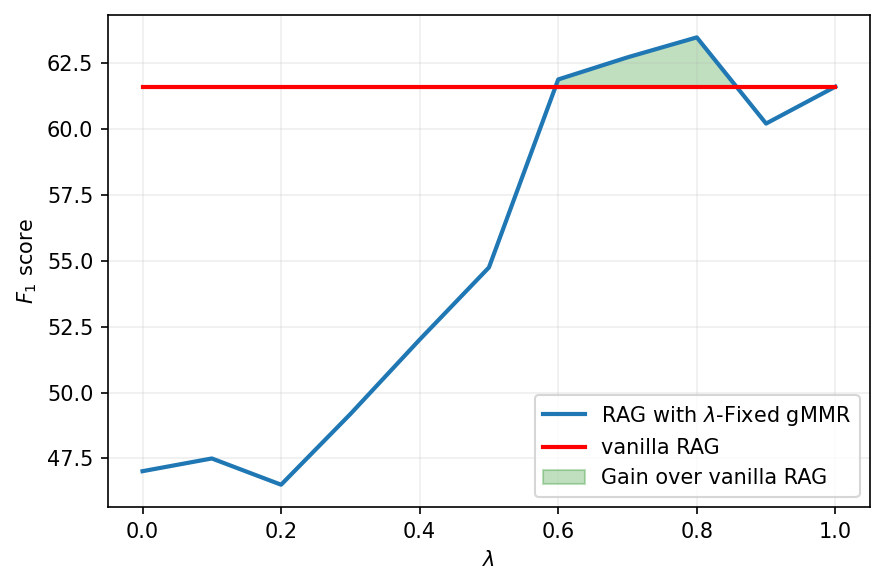}
            \label{fig:plot12}
        \end{subfigure}
    \end{minipage}

    \vspace{0.5cm}

    \parbox[c][4cm][c]{2em}{\rotatebox{90}{\centering \textbf{MultiFieldQA}}}%
    \begin{minipage}[c][4cm][c]{0.9\textwidth}
        \centering
        \begin{subfigure}[b]{0.33\textwidth}
            \centering
            \includegraphics[width=\textwidth]{ 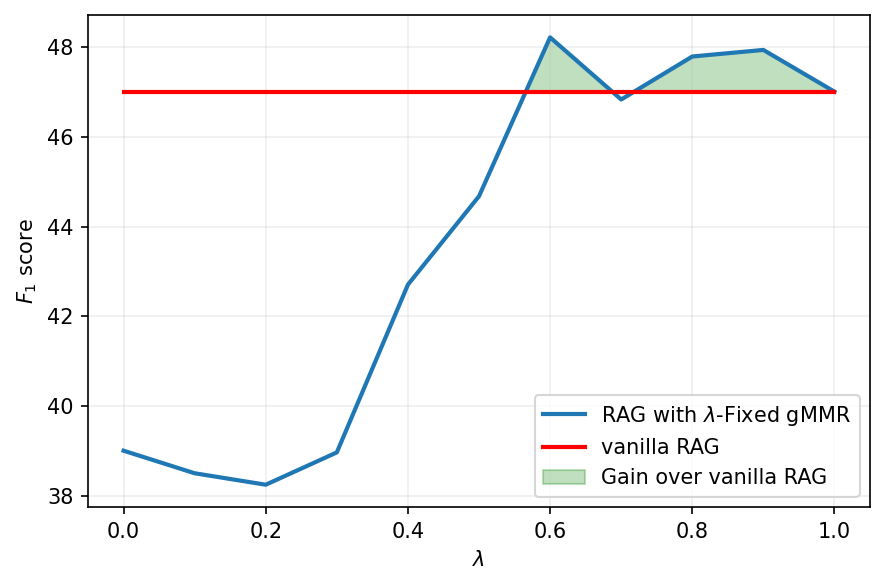}
            \label{fig:plot13}
        \end{subfigure}%
        \hfill%
        \begin{subfigure}[b]{0.33\textwidth}
            \centering
            \includegraphics[width=\textwidth]{ 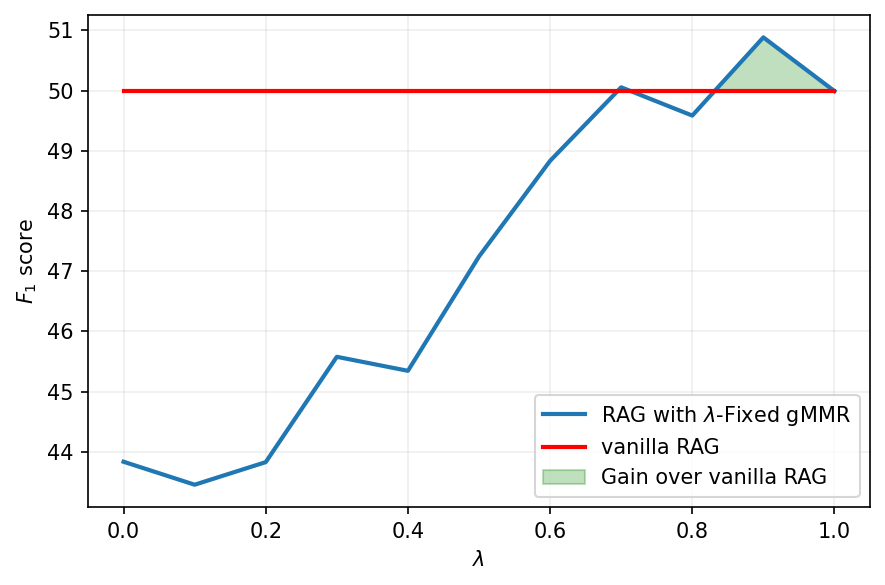}
            \label{fig:plot14}
        \end{subfigure}%
        \hfill%
        \begin{subfigure}[b]{0.33\textwidth}
            \centering
            \includegraphics[width=\textwidth]{ 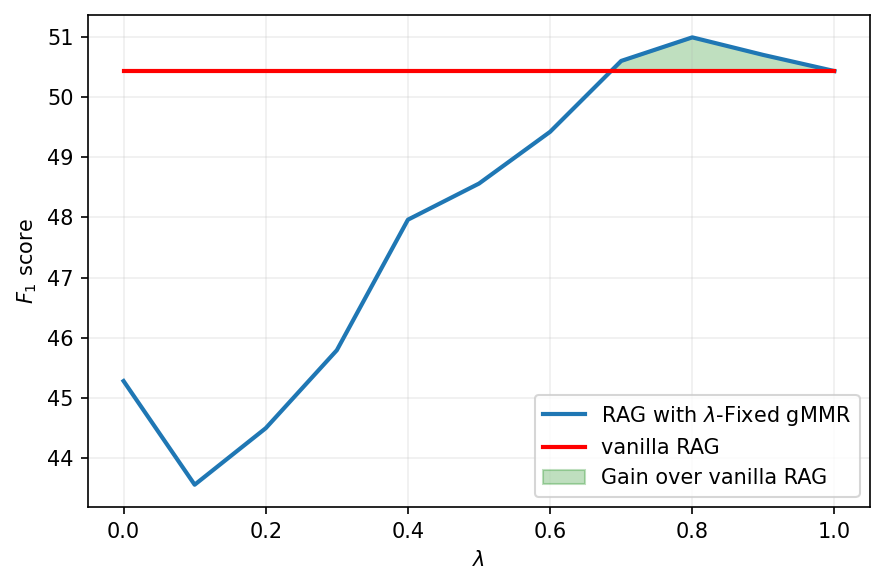}
            \label{fig:plot15}
        \end{subfigure}
    \end{minipage}
    \vspace{0.5cm}
    \parbox[c][4cm][c]{2em}{\rotatebox{90}{\centering \textbf{EN.QA}}}%
    \begin{minipage}[c][4cm][c]{0.9\textwidth}
        \centering
        \begin{subfigure}[b]{0.33\textwidth}
            \centering
            \includegraphics[width=\textwidth]{ 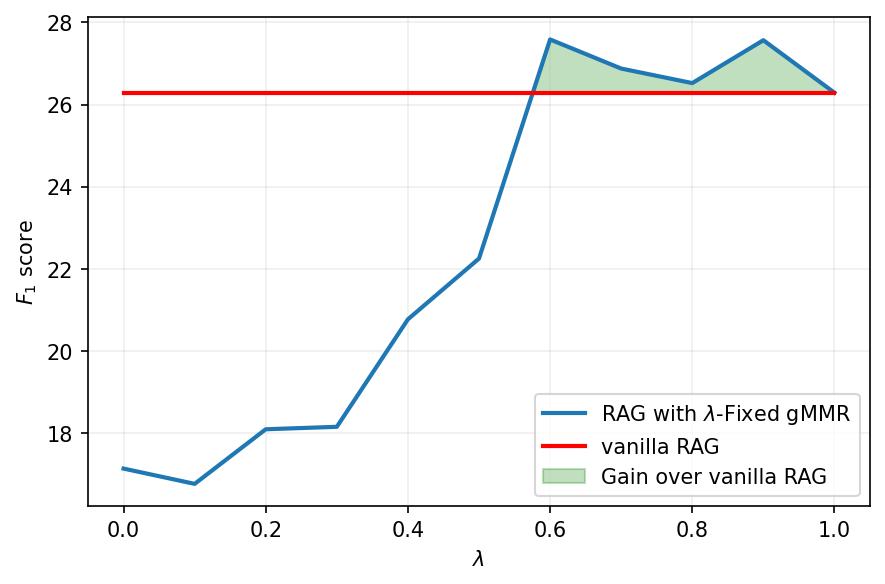}
            \label{fig:plot7}
        \end{subfigure}%
        \hfill%
        \begin{subfigure}[b]{0.33\textwidth}
            \centering
            \includegraphics[width=\textwidth]{ 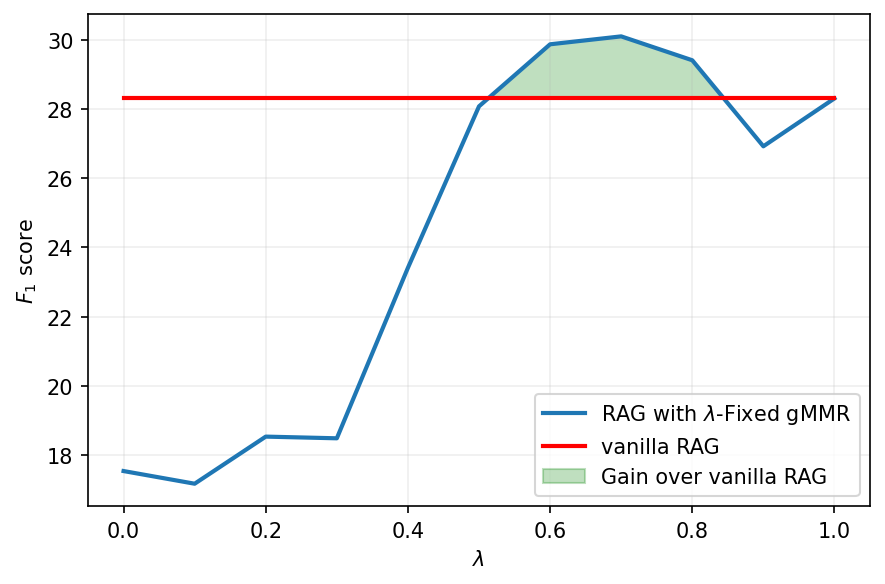}
            \label{fig:plot8}
        \end{subfigure}%
        \hfill%
        \begin{subfigure}[b]{0.33\textwidth}
            \centering
            \includegraphics[width=\textwidth]{ 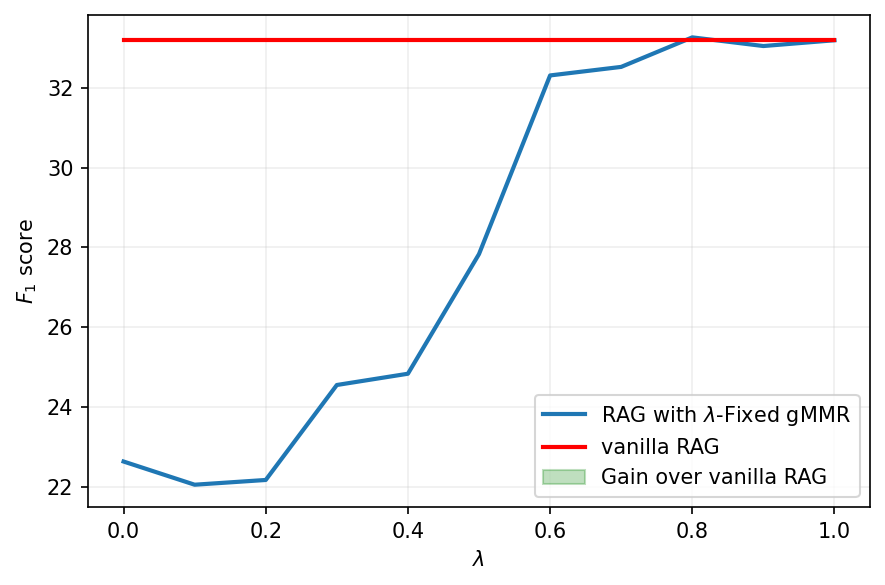}
            \label{fig:plot9}
        \end{subfigure}
    \end{minipage}

    \caption{$F_1$ Scores comparison of RAG with $\lambda$-Fixed gMMR vs vanilla RAG at different chunk sizes for our 5 benchmarks utilizing Llama 3.3 70B as the backbone LLM.}
    \label{fig:RAGwMMR}
\end{figure*}

Figure~\ref{fig:RAGwMMR} presents a detailed performance comparison between vanilla RAG and RAG with $\lambda$-Fixed gMMR across the five benchmarks at the different context lengths settings with Llama 3.3 70B as the backbone LLM. In vanilla RAG, performance is invariant to $\lambda$, producing a horizontal line, whereas RRAG with $\lambda$-Fixed gMMR's performance varies with $\lambda$, with each benchmark exhibiting a clear peak at a given context length. For each benchmark, we find at least one context length contains a $\lambda$ where RAG with $\lambda$-Fixed gMMR outperforms vanilla RAG. At the 500-word context length, the benefits of using RAG with $\lambda$-Fixed gMMR is more prominent. whereas for 1000 and 1500-word context length settings, $F_1$ improvements are less substantial.

\subsection{Performance comparison with classical MMR vs gMMR}
\label{sec:eucvs}

We compare DF-RAG using classical MMR with DF-RAG using our proposed gMMR retrieval function. The only difference in these two retreival functions is the distance metric, where traditional MMR uses cosine similarity and gMMR uses Euclidean distance in a normalized embedding space. Experiments are conducted with 500-word chunks and Llama 3.3 70B as the backbone LLM (see Table \ref{tab:computation}). DF-RAG with gMMR outperforms the classical MMR variant on three out of five benchmarks, and closely matches its performance on the remaining two. Overall, gMMR yields higher average gains compared to classical MMR.
\begin{table}[h]
\centering

\begin{tabular}{lcc}
\toprule
\textbf{Benchmarks} & \textbf{classical MMR} & \textbf{gMMR} \\
\midrule
HotpotQA & 56.7 & \textbf{58.1} \\
MuSiQue  & \textbf{34.1} & 34.0 \\
2WikiMQA      & 53.6 & \textbf{58.1} \\
MultifieldQA       & \textbf{49.0} & 48.3\\
En.QA       & 28.2 & \textbf{29.3} \\
\bottomrule
\end{tabular}%

\caption{$F_1$ scores of DF-RAG performing with classical MMR vs gMMR}
\label{tab:computation}
\end{table}

%
%
%
%
%

\subsection{DF-RAG with Binary-Search sampling}
\label{sec:bs}

DF-RAG’s latency is primarily determined by the number of LLM calls made by the Evaluator, which is in turn contingent upon the sampling strategy used to sample chunk sets. To improve efficiency, we introduce a binary-search–based sampling strategy, motivated by its effectiveness in locating peaks in unimodal distributions. This approach reduces the sampling complexity from $n$ to $\log n$. Binary-search sampling achieves comparable performance to uniform sampling across models (see Table \ref{tab:results_halfpage}), with only minimal drops in accuracy.

\begin{table*}[h]
\centering
\scriptsize
\begin{tabular}{l cc cc}
\toprule
\multirow{2}{*}{\textbf{Dataset}} 
& \multicolumn{2}{c}{\textbf{DF-RAG + IC (uniform sampling)}} 
& \multicolumn{2}{c}{\textbf{DF-RAG + IC (Binary-Search)}} \\
\cmidrule(lr){2-3} \cmidrule(lr){4-5}
& \textbf{Qwen 2.5 72B} & \textbf{Llama 3.3 70B} 
& \textbf{Qwen 2.5 72B} & \textbf{Llama 3.3 70B}  \\
\midrule
HotPotQA & 60.1 & 56.8 & 56.0 & 54.1 \\
MuSiQue & 32.7 & 34.0 & 30.3 & 33.4 \\
2WikiMQA & 56.8 & 55.2 & 54.3 & 50.5 \\

%
%

\bottomrule
\end{tabular}
\caption{Comparison of DF-RAG + IC with uniform sampling and Binary-Search based sampling.}
\label{tab:results_halfpage}
\end{table*}

\section{Error Analysis}
\label{sec:ea}
We analyze certain cases where DF-RAG both succeeds and fails and try to draw intuition about why such cases arise. In these cases, we compare DF-RAG with vanilla RAG to better understand the point of success and failure.\\

\subsection{DF-RAG successful cases}
In most successful cases, we find that DF-RAG is able to retrieve information that vanilla RAG overlooks. This often occurs when DF-RAG captures chunks that align with each reasoning step required to answer a multi-hop query by fetching diverse and complementary information, whereas vanilla RAG often misses key bridging information because such chunks have low surface similarity to the query and are therefore excluded by cosine-similarity–based retrieval. An illustrative example is provided in Table \ref{tab:success}. For the input query \textit{What is the place of birth of the performer of the song Ruleta (Inna Song)?}, DF-RAG retrieves both complementary pieces of information: \textit{Ruleta is recorded by Romanian singer Inna} and \textit{Inna Elena Alexandra Apostoleanu was born in Mangalia} (highled in the example). In contrast, vanilla RAG only retrieves the first piece of information about \textit{Ruleta being recorded by Inna}, but not the birthplace information. Vanilla RAG retrieves many chunks that mostly discuss Ruleta because they are highly relevant to the query, but this displaces the other information needed to make the final reasoning step and generate the answer, i.e., \textit{Inna being born in Mangalia}.\\


\subsection{DF-RAG failure cases}
Our analysis to visualize certain cases where DF-RAG fails revealed that DF-RAG sometiems fails to retrieve information that is needed for reasoning over multi-hop QAs, and this is a limitation common to RAG that stems from the inherent nature of information loss in retrieval. However, there are also cases where key information is retrieved appropriately, but due to the structure of the chunks presented, the Generator finds it difficult to reason effectively to generate the answer (see Table \ref{tab:failure}). For the question \textit{What is the place of birth of the director of the film Ninamaninja Kalpadukal?}, it is important first to find \textit{who the director of Ninamaninja Kalpadukal is} and second \textit{where his birthplace is}. If we look at the chunks retrieved by both DF-RAG and vanilla RAG, we can see that both pieces of information are retrieved by the systems (highlighted in the examples). However, it is important how the information is presented to the generator. For the chunks retrieved by DF-RAG, the chunk that talks about the village \textit{Methala} appears somewhat abruptly, making it difficult to decipher that Methala is the birthplace of the identified director of Ninamaninja Kalpadukal. By contrast, in the chunks retrieved by vanilla RAG, the information required for performing the second reasoning step comes right after introducing the director of Ninamaninja Kalpadukal, making it sound like the place where the director spent his early life, and thus the generator can bridge the gap to generate \textit{Methala} as the answer.


\subsection{Understanding performance trends across benchmarks}
\label{sec:perf}
\textbf{Multi-hop QA benchmarks:} We observe that DF-RAG is able to close a substantial portion of the performance gap between vanilla RAG and our Oracle on two of the three multi-hop QA benchmarks—2WikiMQA and HotpotQA. However, the improvements are smaller on MuSiQue for DF-RAG compared to vanilla RAG. One reason for this discrepancy is the relative complexity of the benchmarks. HotpotQA predominantly contains 2-hop questions, where the reasoning chain involves fewer bridges and is thus easier compared to questions requiring more hops. For instance, in the question: \textit{Prior to playing for Michigan State, Keith Nichol played football for a school located in what city?}, the reasoning needs to identify: 1) \textit{Identify the school Keith Nichol played for before Michigan State} and 2) \textit{Determine the city where that school is located}. Both of these questions can be answered solely by relying on the query as the entities of interested are directly placed there. By retrieving a sufficiently diverse set of information about \textit{Keith Nichol}, we can find details about his school prior to Michigan State. Similarly, by retrieving information about \textit{schools}, we can also identify the city in which the school is located.   

\noindent In contrast, MuSiQue contains a greater number of hops per question. For example, the question \textit{When was the last time Peter Till's sports team beat the winner of the 1894–95 FA Cup?} requires essentially three hops: 1) \textit{Identify Peter Till's sports team}, 2) \textit{Identify the winner of the 1894–95 FA Cup}, and 3) \textit{Determine the last time the team identified in the first step defeated the winner of the 1894–95 FA Cup}. We can answer the first hop by relying on retrieving a sufficiently diverse retrieval set on Peter Till. We can also perform the second hop by retriving information about the winners of FA cup on 1894-1895. However the third hop needs connecting the entities we find in the first and second hop, which is impossible to guess without answering them and therefore retrieving information solely based on question provides minimal support. As a result, unlike in HotpotQA, simply relying on the question often provides limited guidance on what evidence needs to be retrieved to answer the final hop and hence both DF-RAG and Vanilla RAG fails. Moreover, MuSiQue mentions that for each question presented, it is compulsory to perform each hop as they filtered out questions where later hops could be answered without knowing the earlier answers, making no shortcut jumps possible and a difficult dataset to address at short context scenarios. \\

\noindent\textbf{Non multi-hop QA benchmarks:} DF-RAG also demonstrates improvements on MultifieldQA and En.QA. Although these datasets are not inherently multi-hop, their contexts often contain distracting information that appears relevant but leads models to incorrect answers. DF-RAG mitigates this by retrieving not only relevant but also diverse evidence, increasing the likelihood of capturing the true answer amid such noise. For example, in MultifieldQA, the question \textit{What is the score achieved by the authors for Track-2?} often misleads vanilla RAG. The retriever selects a chunk stating \textit{All of our submissions were the top submissions for each track, which surpassed the next best competitors by a margin of 4.5\% and 5.6\% for Track-1 and Track-2 respectively}, causing the model to output 5.6\%, which is the margin of improvement rather than the score. In contrast, DF-RAG successfully retrieves the chunk \textit{We achieve a score of 58.54\% for Track-1 and 85.61\% for Track-2}, which, despite surrounded with noisy information, provides the correct answer.

\begin{table*}[h!]
\centering
\renewcommand{\arraystretch}{1.3} 
\begin{tabular}{p{\linewidth}}
\hline
\textbf{Question:} What is the place of birth of the performer of song Ruleta (Inna Song)? \\
\hline
\textbf{Selected chunks by DF-RAG}:
EDM and reggaeton-influenced track, containing touches of Indian and Caribbean music. When asked by a Direct Lyrics interviewer to describe the lyrics and sound of "Ruleta", Inna said, "It's summerish, it's cool, it's fun and I dare to say... it's catchy and addictive!" Reception Upon its release, "Ruleta" received positive reviews from music critics. Currinn from CelebMix wrote, "this new song has Inna proving that she's multilingual", and compared the style of "Ruleta" to her "Heaven" (2016). Kevin Apaza, writing for Direct Lyrics thought, "summer doesn't officially start until Romanian pop queen Inna releases her summer single", further calling it
Passage 1: \hl{Inna Elena Alexandra Apostoleanu (born 16 October 1986), known professionally as Inna (stylized in all caps), is a Romanian singer and songwriter. Born in Mangalia and raised in Neptun, she studied political science at Ovidius University before meeting the Romanian trio Play \& Win and pursuing a music career.} She adopted the stage name "Alessandra" and a pop-rock style in 2008; later that year, she changed her stage name to "Inna" and began releasing house music. "Hot" (2008), her debut single, was a commercial success worldwide and topped the Romanian and Billboard's Hot Dance Airplay chart, among others.
mtvU for his video that depicted a love affair between a woman and her television, with the TV containing the green-screened Decemberists video footage. Passage 3: \hl{Ruleta (Inna song) "Ruleta" (English: "Roulette") is a song recorded by Romanian singer Inna}, featuring collaborative vocals by Romanian performer Erik, for her fifth studio album, Nirvana (2017) ... \\

\hline

\textbf{Selected Chunks by Vanilla RAG:} ... An editor of Bihoreanul praised the "attractive" rhythms of "Ruleta" and noted its potential commercial success in Romania.Commercially, the song debuted at number 32 on Romania's Airplay 100 for the week ending 9 July 2017 as the highest new entry, marking one of Inna's highest debuts on the chart. In late August 2017, "Ruleta" reached its peak position at number three, becoming the singer's highest charting single since "Bop Bop" (2015), which peaked at number two. The track also opened at number 39 on the French Club 40 chart for the week ending 15 July 2017 and
rose up to number 26 on the following week. "Ruleta" further charted in other countries, most notably reaching number two in Bulgaria. Music video An accompanying music video for "Ruleta" was shot at the Domeniul Greaca resort in Romania by Barna Nemethi, with Marius Apopei serving as the director of photography. Inna recalled the experience, "Actually, I was super chill in that day, super relaxed, just going with the flow. Most of the times, I'm kind of stressed out, because I want everything to be perfect. Not too many funny stuff, just having fun."The clip was uploaded onto her official
mtvU for his video that depicted a love affair between a woman and her television, with the TV containing the green-screened Decemberists video footage. Passage 3: \hl{Ruleta (Inna song) "Ruleta" (English: "Roulette") is a song recorded by Romanian singer Inna, featuring collaborative vocals by Romanian performer Erik, for her fifth studio album, Nirvana (2017)}. It was made available for digital download and streaming on 21 June 2017 by Global Records as the album's second single. The track was written by Inna, Erik, Marius Dia and Breyan Isaac, while production was handled by Sebastian Barac, Marcel Botezan and David Ciente. \\
\hline

\textbf{Answer of DF-RAG:} Mangalia \\

\textbf{Vanilla RAG Answer:} Romania \\
\textbf{Golden Answer:} Mangalia \\
\hline
\end{tabular}
\caption{Selected chunks of a successful case by DF-RAG and Vanilla RAG. DF-RAG successfully retrieves the information needed to perform the 2nd hop to answer the query.}
\label{tab:success}
\end{table*}

\begin{table*}[t]
\centering
\renewcommand{\arraystretch}{1.3} 
\begin{tabular}{p{\linewidth}}
\hline
\textbf{Question:} What is the place of birth of the director of film Ninamaninja Kalpadukal? \\
\hline
\textbf{Selected chunks by DF-RAG}:
an assistant director for Tamil - Telugu movies in Newton Studios, he later distinguished himself as a producer, director and scriptwriter. His directorial debut, Ninamaninja Kalpadukal, which portrays the trials the Indo-China war, won 4 awards including the President's silver medal for best regional film, and the award for The Best Director. He directed about 6 movies and produced 2 movies on his own. He wrote the screen play and directed 4 episodes of Aithihyamala for Doordarshan. He also directed a tele-film titled Kudajadri. Filmography Passage 4: Peter Levin Peter Levin is an American director of film, television and theatre.
\hl{into a family known as "Kallil" in Methala, Near Kalady in Kerala, he completed his schooling in Paravoor and then was graduated in economics from Serhampur}, Bengal. He was an avid reader always and appreciated the nuances of literature. His first short story was published in the weekly Prasanna Keralam from Kottayam when he was studying in high school. Since then, his stories have appeared in weeklies. Writing career One of his first of his short stories was published in the weekly Prasanna Keralam from Kottayam, when he was studying in high school.The editor of Kaumudi, K Balakrishnan, lead him
Norwegian Museum of Cultural History. In 2010 he was decorated with the Royal Norwegian Order of St. Olav. Passage 8: S. N. Mathur S.N. Mathur was the Director of the Indian Intelligence Bureau between September 1975 and February 1980. He was also the Director General of Police in Punjab. Passage 9: Ian Barry (director) Ian Barry is an Australian director of film and TV. Select credits Waiting for Lucas (1973) (short) Stone (1974) (editor only) The Chain Reaction (1980) Whose Baby? (1986) (mini-series) Minnamurra (1989) Bodysurfer (1989) (mini-series) Ring of Scorpio (1990) (mini-series) Crimebroker (1993) Inferno (1998) (TV movie) Miss
Passage 1: \hl{Ninamaninja Kalpadukal Ninamaninja Kalpadukal (Bloodstained Footprints) is a 1963 Malayalam language film, directed by N. N. Pisharody and produced by N.K. Karunakara Pillai and Shobhana Parameswaran Nair}. The lead role is played by Prem Nazir, with Ambika, Sheela and Madhu, who debuted with this film. The film is based on a novel by Parappurath and portrays the trials the Indo-China war. It won the National Film Award for Best Feature Film in Malayalam. It was a super hit movie.There are many evergreen songs in the film, including ... \\ 

\hline

\textbf{Selected Chunks by Vanilla RAG:} an assistant director for Tamil - Telugu movies in Newton Studios, he later distinguished himself as a producer, director and scriptwriter. His directorial debut, Ninamaninja Kalpadukal, which portrays the trials the Indo-China war, won 4 awards including the President's silver medal for best regional film, and the award for The Best Director. He directed about 6 movies and produced 2 movies on his own. He wrote the screen play and directed 4 episodes of Aithihyamala for Doordarshan. He also directed a tele-film titled Kudajadri. Filmography Passage 4: Peter Levin Peter Levin is an American director of film, television and theatre.
Passage 1: \hl{Ninamaninja Kalpadukal Ninamaninja Kalpadukal (Bloodstained Footprints) is a 1963 Malayalam language film, directed by N. N. Pisharody and produced by N.K. Karunakara Pillai and Shobhana Parameswaran Nair}. The lead role is played by Prem Nazir, with Ambika, Sheela and Madhu, who debuted with this film. The film is based on a novel by Parappurath and portrays the trials the Indo-China war. It won the National Film Award for Best Feature Film in Malayalam. It was a super hit movie.There are many evergreen songs in the film, including "Mamalakalkkappurathu" (by P. B. Sreenivas) and "Anuraga Natakathil" (by K. P.
\hl{into a family known as "Kallil" in Methala, Near Kalady in Kerala, he completed his schooling in Paravoor and then was graduated in economics from Serhampur}, Bengal. He was an avid reader always and appreciated the nuances of literature. His first short story was published in the weekly Prasanna Keralam from Kottayam when he was studying in high school. Since then, his stories have appeared in weeklies...\\ 

\hline

\textbf{Answer of DF-RAG:} not specified in the provided passages. \\

\textbf{Vanilla RAG Answer:} methala, near kalady in kerala \\
\textbf{Golden Answer:} Methala \\
\hline
\end{tabular}
\caption{Selected chunks of a failure case by DF-RAG and Vanilla RAG. DF-RAG successfully retrieves the information needed but fails to stitch the reasoning gap to answer the query.}
\label{tab:failure}
\end{table*}



\begin{figure*}[t]
\begin{tcolorbox}[
    colback=white,
    colframe=gray,
    boxrule=1pt,
    arc=0pt,
    width=\textwidth,
    title={\textbf{Planner}},
    fonttitle=\normalsize
]
Instruction: You are an expert question decomposer planner. \\[6pt]
Task: Break the user's \textbf{single question} into the minimal
        ordered list of sub-questions that must all be answered
        to fully answer the original question harnessing your zero-shot capabilites. 
        Follow the output format exactly as shown in the example below to generate a plan (a list of sub-questions).

        \vspace{0.3cm}
        
        Question: Where was the author of the book `Jacob' Born?
        
        \vspace{0.3cm}
        Output Planner :\\
        1) Identify the author of Jacob\\
        2) Identify where the author was born
        \vspace{0.3cm}
        
        Only produce the PLAN section — no extra commentary for the following question:
        
        \vspace{0.3cm}
        Question: \textcolor{blue}{<Question>}
\end{tcolorbox}
\caption{Prompt used by the Planner}
\label{fig:planner}
\end{figure*}

\begin{figure*}[t]
\begin{tcolorbox}[
    colback=white,
    colframe=gray,
    boxrule=1pt,
    arc=0pt,
    width=\textwidth,
    title={\textbf{Evaluator}},
    fonttitle=\normalsize
]
Instruction: You are a strict judge evaluating text chunks. Only count information that is explicitly stated. \\[6pt]

Task: Score each question 0-5 based on how directly it can be answered:\\
5: EXPLICIT answer with specific facts/numbers stated directly\\
4: Clear and complete answer with direct relevant information present\\ 
3: Mostly complete answer, minor missing detail, but can be inferred\\
2: Partial information, key details missing\\
1: Very limited relevance, mostly vague \\
0: No relevant information or requires inference\\

STRICT RULES: \\ 
- Only count information DIRECTLY STATED in chunks \\ 
- Do NOT give points for logical leaps or inferences \\
- Do NOT give points for vague or tangential content \\

Your reasoning for the score values you provide should be within compact. **You MUST remember this.** 
At the end you must answer in the format: Total Score: <Total Score>  \\
Follow the examples for understanding the task: \\

\textcolor{red}{<Few Shot Examples>} \\

Now do the same for the following: \\
Plan: \textcolor{red}{<Plan>}

Chunks:
\textcolor{red}{<Candidate Chunk Set>} \\
\end{tcolorbox}
\caption{Prompt used by the Evaluator}
\label{fig:exec}
\end{figure*}

\begin{figure*}[t]
\begin{tcolorbox}[
    colback=white,
    colframe=gray,
    boxrule=1pt,
    arc=0pt,
    width=\textwidth,
    title={\textbf{Few Shot examples for Evaluator}},
    fonttitle=\normalsize
]

    --------------------------------------------------------------------------------------------------------------------
   \\
    Example 1
    \\
    --------------------------------------------------------------------------------------------------------------------
    \\
    Plan:\\
    1) Find the protagonist in the movie 'Inception'\\
    2) Find the protagonist's birthplace\\
    3) Find the population of that city\\
    
    Chunks:\\
    "Inception is a 2010 science fiction film starring Leonardo DiCaprio as Dom Cobb, a professional thief who infiltrates people's dreams..."
    "Leonardo DiCaprio was born in Los Angeles, California, United States on November 11, 1974..."
    "Los Angeles is the most populous city in California with a population of approximately 4 million people as of 2023..."\\
    
    For each question in the plan, judge if the information needed is clearly and explicitly present in the chunks. The chunk set should collectively cover all different types of information required by the steps.\\
    
    1. Score: 5. Short Explanation: Leonardo DiCaprio is explicitly named as the protagonist.\\
    2. Score: 5. Short Explanation: Born in Los Angeles, California is directly stated.\\
    3. Score: 5. Short Explanation: Population of 4 million is explicitly mentioned.\\
    Total Score: 15\\
    
   --------------------------------------------------------------------------------------------------------------------
   \\
    Example 2
    \\
    --------------------------------------------------------------------------------------------------------------------
    \\
    Plan:\\
    1) Find the director of the movie 'Titanic'\\
    2) Find the director's net worth\\
    3) Find the director's upcoming projects\\
    
    Chunks:\\
    "Titanic is a 1997 epic romance and disaster film. The movie was a massive box office success..."
    "The film won 11 Academy Awards including Best Picture and Best Director..."
    "Leonardo DiCaprio and Kate Winslet starred as the main characters in this epic love story..."\\
    
    For each question in the plan, judge if the information needed is clearly and explicitly present in the chunks. The chunk set should collectively cover all different types of information required by the steps.\\
    
    1. Score: 0. Short Explanation: Director name not mentioned, only that it won Best Director\\
    2. Score: 0. Short Explanation: No financial information about anyone provided\\
    3. Score: 0. Short Explanation: No information about future projects mentioned\\
    Total Score: 0\\

\end{tcolorbox}
\caption{Few Shot Examples used by the Evaluator}
\label{fig:few-shot}
\end{figure*}

\begin{figure*}[t]
\begin{tcolorbox}[
    colback=white,
    colframe=gray,
    boxrule=1pt,
    arc=0pt,
    width=\textwidth,
    title={\textbf{Few Shot Examples contd.}},
    fonttitle=\normalsize
]
-------------------------------------------------------------------------------------------------------------------
   \\
    Example 3
    \\
    --------------------------------------------------------------------------------------------------------------------
    \\
    
    Plan:\\
    1) Find the author of 'Harry Potter'\\
    2) Find the author's age\\
    3) Find the number of books in the series\\
    
    Chunks:
    "Harry potter, the book written by the sister of James Franco is fantasy book series that became globally popular..."
    " James Franco and along with all his siblings travelled to Europe..."
    "In Europe, Franco and JK Rowling played Hokkey like all the siblings"
    "The series consists of seven main novels plus several companion books..."
    "Rowling has become one of the most successful authors in modern publishing..."\\
    
    For each question in the plan, judge if the information needed is clearly and explicitly present in the chunks. The chunk set should collectively cover all different types of information required by the steps.
    
    1. Score: 3. Short Explanation: J.K. Rowling is inferred named as the author from the relationships.\\
    2. Score: 0. Short Explanation: Age not mentioned anywhere in the chunks.\\
    3. Score: 5. Short Explanation: Seven main novels is directly stated.\\
    Total Score: 8\\

\end{tcolorbox}
\caption{Few Shot Examples used by the Evaluator contd.}
\label{fig:few-shot-contd}
\end{figure*}

\begin{figure*}[t]
\begin{tcolorbox}[
    colback=white,
    colframe=gray,
    boxrule=1pt,
    arc=0pt,
    width=\textwidth,
    title={\textbf{Generator}},
    fonttitle=\normalsize
]
Instruction: You are a question answering assistant. \\[6pt]
Task:
Answer the question based on
the given passages. Only give
me the answer and do not output
any other words. The following
are given passages. 
\textcolor{red}{<Context>}

Answer the question based on
the given passages. Only give
me the answer and do not output
any other words. 

Question: \textcolor{red}{<Query>}

Answer:

\end{tcolorbox}
\caption{Prompt used by the Generator}
\label{fig: gen-prompt}
\end{figure*}


\end{document}